%% file: main.tex
\documentclass[pdflatex,sn-mathphys-num]{sn-jnl}% Math and Physical Sciences Numbered Reference Style
\usepackage[utf8]{inputenc} % allow utf-8 input
\usepackage[T1]{fontenc}    % use 8-bit T1 fonts
\usepackage{lmodern}        % scalable fonts for microtype font expansion
\usepackage{hyperref}       % hyperlinks
\usepackage{url}            % simple URL typesetting
\usepackage{booktabs}       % professional-quality tables
\usepackage{amsmath}        % math
\usepackage{amsfonts}       % blackboard math symbols
\usepackage{amssymb}        % math symbols
\usepackage{nicefrac}       % compact symbols for 1/2, etc.
\usepackage{microtype}      % microtypography
\usepackage{xcolor}         % colors
\usepackage{graphicx}       % figures
\usepackage{multirow}       % tables
\usepackage{array}          % tables
\usepackage{wrapfig}
\usepackage{compact-layout}
\hypersetup{bookmarksdepth=3} % include section levels, omit run-in paragraph bookmarks
\theoremstyle{thmstyleone}%
\theoremstyle{thmstyletwo}%

\theoremstyle{thmstylethree}%

\begin{document}

\title[Multimodal Fusion of RGB and Complementary Modalities for Semantic Segmentation]{Multimodal Fusion of RGB and Complementary Modalities for Semantic Segmentation}

\author[1]{\fnm{Hu} \sur{Cao}}

\author[2]{\fnm{Yifeng} \sur{Cheng}}

\author[2]{\fnm{Mu} \sur{He}}

\author[2]{\fnm{Mengyu} \sur{Li}}

\author[2]{\fnm{Yinlong} \sur{Liu}}

\author[2]{\fnm{Alois} \sur{Knoll}}

\affil[1]{\orgdiv{School of Automation}, \orgname{Southeast University}, \orgaddress{\street{Sipailou 2}, \city{Nanjing}, \postcode{210096}, \country{China}}}
\affil[2]{\orgdiv{Chair of Robotics, Artificial Intelligence and Real-time Systems}, \orgname{Technical University of Munich}, \orgaddress{\street{Boltzmannstraße 15}, \city{Munich}, \postcode{85748}, \country{Germany}}}


\abstract{
Multi-modal semantic segmentation adds an auxiliary sensing stream to RGB images (RGB-X), such as thermal, LiDAR, depth, event, polarization, or light field. This can improve robustness in poor lighting and under motion blur. But different sensors often produce features that do not match well, and fusion can become unstable. 
To alleviate these issues, we propose a bidirectional polarity-aware cross-modality fusion (BPCF) module that effectively captures complementary cues while enhancing feature alignment. We evaluate the framework on six modality pairings: RGB-Thermal, RGB-LiDAR, RGB-Depth, RGB-Event, RGB-Polarization, and RGB-Light Field. Our model achieve state-of-the-art (SOTA) results on eight public datasets, including  MFNet, KITTI-360, DELIVER, DDD17, DSEC, MCubeS, ZJU, and UrbanLF. 
The codes and trained models will be released upon acceptance.
}




\maketitle

\input{sections/01_introduction}

\input{sections/02_related_work}
\input{sections/03_method}
\input{sections/04_experiments}

\input{sections/06_discussion}
\input{sections/05_conclusion}

\section*{Funding Information}
This work was supported by the National Natural Science Foundation of China (Grant No. 62603145). The funding body had no role in the design of the study; in the collection, analysis, or interpretation of data; in the writing of the manuscript; or in the decision to publish the results.

\section*{Conflict of Interest Statement}
The authors declare that they have no conflict of interest regarding the publication of this paper.

\section*{Data Availability}
This study uses publicly available datasets. No new data were created or analyzed in this study. Data sharing is not applicable to this article.

\bibliography{main}

\end{document}

%% file: sections/01_introduction.tex
\section{Introduction}

In autonomous driving and robotic perception, RGB-only semantic segmentation often suffers from performance degradation under low illumination, fast motion, and adverse weather conditions~\cite{Bijelic2019SeeingTF,zhang2023delivering}. Fortunately, non-RGB modalities provide complementary cues that can mitigate these limitations. For instance, incorporating thermal, LiDAR, depth, event, light field, and polarization modality data can enhance robustness against low-light conditions, motion blur, and adverse weather conditions~\cite{Chen2020EventBasedNV,Bijelic2019SeeingTF}. Nevertheless, achieving reliable multi-modal fusion that scales to many sensors and consistently leverages each modality's strengths remains an open challenge.

The problem can be considered from two perspectives. 
The first is the efficient utilization of complementary information. In multi-modal scenarios such as RGB–Thermal or RGB–Event, the same object may exhibit reversed attention responses across modalities. 
Consequently, cross-modal interactions in RGB-X semantic segmentation are not purely positively correlated. Depending on the sensor physics and scene conditions, the same semantic structure can elicit aligned responses across modalities or reversed responses. Importantly, these negative correlations are not necessarily noise. Our key observation is that these contrastive responses can be indicative: they can reveal complementary cues that are hard to capture while the fusion mechanism assumes only non-negative similarity.
Recognizing that such negative correlations can also provide valuable cues, we propose a bidirectional polarity-aware linear cross-attention (BPLCA) mechanism. BPLCA decomposes features into positive and negative components and performs cross-modal attention across both branches. This design enables the model to capture not only shared but also complementary signals across modalities, leading to more complete and balanced feature integration.

\begin{wrapfigure}{r}{0.45\textwidth}
    \centering
    \includegraphics[width=\linewidth]{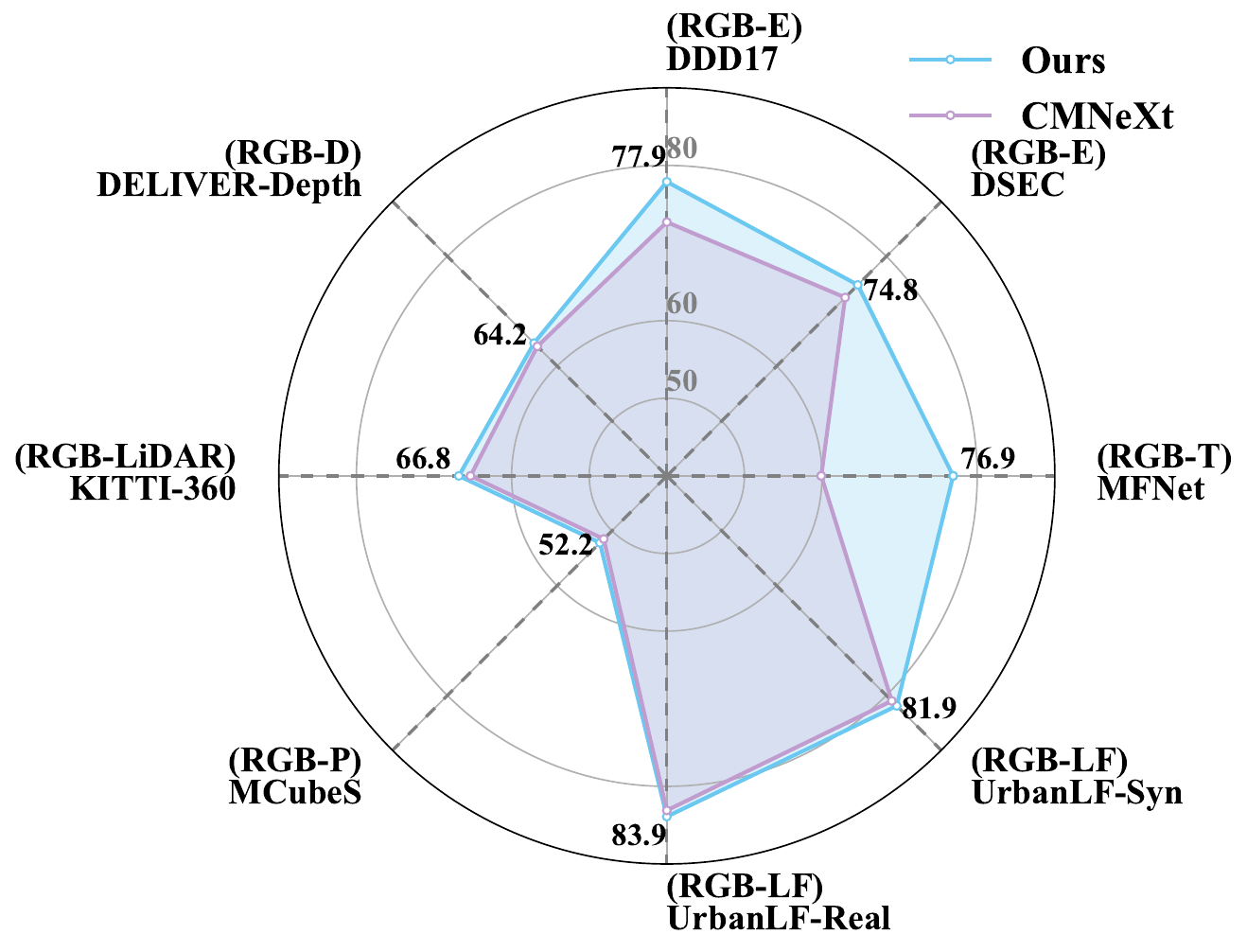}
    \caption{Performance comparison between our proposed method (Ours) and CMNeXt~\cite{zhang2023delivering} across different modality combinations, including RGB-Thermal (MFNet), RGB-LiDAR (KITTI360), RGB-Depth (DELIVER), RGB-Event (DSEC and DDD17), RGB-Polarization (MCubeS), and RGB-Light Field (UrbanLF-Real and UrbanLF-Syn).}
    \label{fig:overview}

\end{wrapfigure}

A second challenge in RGB-X semantic segmentation is cross-modal feature alignment. Heterogeneous modalities (e.g., RGB, thermal, LiDAR, depth, events, and polarization) differ substantially in their statistics and appearance, making naive fusion prone to inconsistent representations or over-reliance on a single modality~\cite{zhang2023cmx,zhang2023delivering,cao2024embracing}. To address this, we introduce a Dual Feature Consistency Constraint (DFCC) that explicitly encourages cross-modal features to become semantically coherent, thereby narrowing the modality gap and stabilizing fusion.
DFCC aligns RGB and auxiliary features by first reducing distribution mismatch, and then applying scene-adaptive affine modulation whose parameters are predicted from fused features. During fusion, it further regularizes cross-modal semantics by constructing spatial attention maps at each stage and penalizing their discrepancies, encouraging consistent attention and representations while retaining complementary cues.

% \wrapfinish

With BPLCA and DFCC integrated into our framework (together with the proposed designs above), we achieve consistent improvements and outperform recent SOTA methods on eight multi-modal datasets (Fig.~\ref{fig:overview}).
The main contributions of this work are as follows:

\begin{itemize}
    \item We propose a multi-modal semantic segmentation framework that supports flexible modality combinations. It enables effective representation learning across diverse RGB-X configurations, including RGB-Thermal, RGB-LiDAR, RGB-Depth, RGB-Event, RGB-Polarization, and RGB-Light Field.
    \item  We develop a bidirectional polarity-aware cross-modality fusion (BPCF) module, which integrates bidirectional polarity-aware linear cross-attention (BPLCA) with a dual feature consistency constraint (DFCC) to effectively fuse heterogeneous modalities. 
    
    \item We introduce a stage-wise constraint loss that progressively enforces consistent cross-modal feature alignment.
    \item We conduct extensive experiments on eight multi-modal datasets covering six auxiliary modalities. The results show that our method consistently outperforms SOTA approaches across all datasets.
\end{itemize}

% \begin{figure}[t]
%   \centering
%   \begin{minipage}{0.48\textwidth} % 左半边
%     \centering
%     \includegraphics[width=\linewidth]{figures/radar_v5.pdf}
%     \caption[Overview of our performance in different dataset]{The radar chart illustrates the mIoU scores achieved by our proposed method ("Ours") versus the CMNeXt~\cite{zhang2023delivering} on various data combinations, including RGB–Event (DELIVER-Event~\cite{zhang2023delivering}, DSEC~\cite{gehrig2021dsec}, DDD17~\cite{Binas2017DDD17ED}), RGB–Thermal (MFNet~\cite{ha2017mfnet}), RGB–Light Field (UrbanLF-Syn, UrbanLF-Real~\cite{sheng2022urbanlf}), RGB–LiDAR (KITTI-360~\cite{liao2022kitti}), and RGB–Polarization (MCubeS~\cite{liang2022multimodal}).}
%     \label{fig:overview}
%   \end{minipage}
%   \hfill
%   \begin{minipage}{0.48\textwidth} 
%     % \includegraphics[...] 
%   \end{minipage}
% \end{figure}

%% file: sections/02_related_work.tex
\section{Related Work}

\subsection{Semantic segmentation}
Semantic segmentation has progressed from CNN-based architectures such as FCN~\cite{Shelhamer2014FullyCN}, U-Net~\cite{Ronneberger2015UNetCN}, and SegNet~\cite{Badrinarayanan2015SegNetAD}, to encoder–decoder designs like DeepLabv3+~\cite{Chen2018EncoderDecoderWA}. More recently, Transformer architectures have become dominant, with models including ViT~\cite{Dosovitskiy2020AnII}, Swin Transformer~\cite{Liu2021SwinTH}, SegFormer~\cite{xie2021segformer}, SDPT~\cite{Cao2024SDPTSD}, serving as fundamental baselines for RGB-only dense prediction. Despite this progress, performance degrades under motion blur, low light, or rain/snow. This motivates multi-modal RGB+X architectures for robust scene understanding, where X denotes auxiliary modalities such as thermal, LiDAR, depth, event cameras, polarization, and light field.

\subsection{Multi-modal semantic segmentation} 
% Despite the progress of pure RGB semantic segmentation~\cite{Shelhamer2014FullyCN,Ronneberger2015UNetCN,Badrinarayanan2015SegNetAD,Chen2018EncoderDecoderWA,Dosovitskiy2020AnII,Liu2021SwinTH,xie2021segformer,Cao2024SDPTSD}, 
% performance still degrades under motion blur, low light, or rain/snow. This motivates multi-modal RGB+X architectures for robust scene understanding, where X denotes auxiliary modalities such as thermal, LiDAR, depth, event cameras, polarization, and light field. 
Prior work recognizes the complementary value of non-RGB modalities for semantic segmentation. It explores fusion methods from pairwise RGB–X integration to modality-agnostic frameworks that unify multiple sensor types. For RGB+event combination, ~\cite{Sun2021EventBasedFF,Zhou2022RGBEventFF,xie2024eisnet, 9787342, zheng2024learning, li2025efficient} integrate events over time and use event-count or activity-rate cues, then use dual-branch backbones together with bidirectional cross-attention, to synchronize features across modalities and scales. 
% Beyond RGB-event fusion, other auxiliary modalities have also been explored~\cite{Prakash2021MultiModalFT, Joze2019MMTMMT, Hazirbas2016FuseNetID}. 
Several studies also evaluate fusion with auxiliary modalities beyond events. Their results indicate that non-RGB modalities provide powerful complementary cues for scene understanding~\cite{Prakash2021MultiModalFT, Joze2019MMTMMT, Hazirbas2016FuseNetID}. KTBNet~\cite{cai2025keep} proposes a parameter-efficient symmetric framework that balances the contributions between RGB and an additional modality such as thermal, thereby preventing the dominance of a single branch. Moreover, better-known works on RGB–P segmentation has shown only limited performance with these dual branch architecture~\cite{liang2022multimodal, zhang2023delivering}. A possible reason is that polarization cues can be inconsistent with RGB appearance, so cross-modal evidence is not always positively correlated and may even be negatively correlated.
Recently, PolaFormer~\cite{Meng2025PolaFormerPL} introduces a learnable mechanism with power-based scaling to explicitly model both positive and negative query-key interactions. PolaFormer targets single-modality linear self-attention, so its polarity design mainly repairs attention expressiveness by recovering missing negative interactions. In contrast, our BPLCA/PLCA targets cross-modality fusion, so its polarity design mainly structures inter-modal evidence by separating shared cues from complementary cues to improve fusion. 
%(including negatively correlated) In parallel,  RoadFormer~\cite{Huang2024RoadFormerDR} integrates RGB and polarization via a dual-branch fusion block and achieves commendable performance.
Nevertheless, the above designs lack modality-agnostic generalization, as they are tied to fixed modality combinations. CMX~\cite{zhang2023cmx} employs cross-modal feature rectification and fusion modules for long-range context exchange across modalities. Its successor CMNeXt~\cite{zhang2023delivering} extends scalability through a self-query hub and a parallel pooling mixer, maintaining compactness while adapting to diverse sensor types. Similarly, Zheng et al.~\cite{Zheng2024CenteringTV} propose a modality-agnostic pipeline that dynamically selects the most informative modalities, while their follow-up~\cite{zheng2024learning} introduces a modality-agnostic Feature Fusion (MFF) module that synthesizes heterogeneous sensor streams into a unified representation. GeminiFusion~\cite{Jia2024GeminiFusionEP} further refines cross-modal interactions using pixel-wise intra- and inter-modal attention, and OmniVec2~\cite{Srivastava2024OmniVec2A} explores a large-scale shared representation space for multi-modal and multi-task learning. 
Although these works have advanced the field, challenges still remain in fully exploiting complementary information across modalities and in achieving consistent feature alignment. In this work, we propose a bidirectional polarity-aware cross-modality fusion (BPCF) module to alleviate these issues.

%% file: sections/03_method.tex
\begin{figure}[tbp]%
  \centering
  \includegraphics[width=0.86\textwidth]{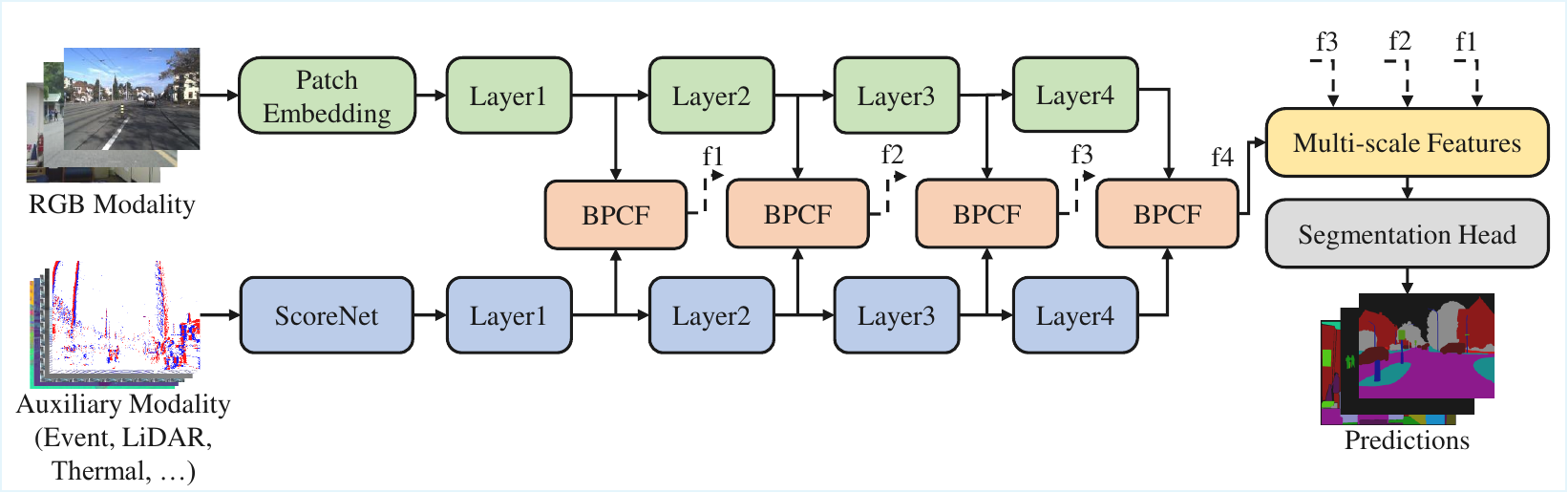}
  \caption[Overview of our proposed model]{Overview of the proposed multi-modal semantic segmentation framework. Each modality is independently encoded using a dedicated Transformer backbone. RGB and auxiliary modality features are then progressively fused through the proposed fusion module in a stage-by-stage manner. Finally, the fused multi-scale features are passed to the segmentation head to generate predictions.}
  \label{fig:full}

\end{figure}

% \begin{multicols}{2}

\section{Method}
\label{sec:method}

As shown in Fig.~\ref{fig:full}, our model adopts an encoder–decoder architecture with stage-wise fusion modules. Each modality is first encoded by a dedicated Transformer backbone~\cite{xie2021segformer}. Following prior work~\cite{zhang2023cmx, zhang2023delivering}, we incorporate ScoreNet to flexibly integrate RGB with one or more auxiliary modalities. For multiple auxiliary inputs, ScoreNet selects the most informative feature at each spatial location. The stage-wise BPCF fusion modules then combine the hierarchical feature representations from the two modality branches. Finally, the fused multi-scale features are aggregated by an MLP-based segmentation head to produce final predictions.

% \columnbreak
% \noindent
  
% \end{multicols}

\subsection{Feature Extraction}

\subsubsection{ScoreNet} 
To extract the most informative modality features, we employ ScoreNet with a dynamic selection mechanism. Specifically, ScoreNet processes each patch in each auxiliary modality as follows:

\begin{equation}
    \begin{aligned}
    S_{i,j} = \text{Softmax}(\text{Linear}(
    \text{GELU}(\text{Linear}(\text{Norm}(X_{i,j}))))),
    \end{aligned}
\end{equation}

for $i \in [1, M]$ and $j \in [1, N]$, where $S_{i,j}$ denotes the informative score for the $i$-th input modality on the $j$-th patch token, $M$ refers to the number of modalities and $N=H\times W$ refers to the number of patches. 
For each patch token, the embedding from the modality with the highest score is dynamically selected and fused into a single auxiliary input to complement the primary RGB data:
\begin{equation}
    x_{\text{a}}^{(0)}=\{X_{m, j}|j\in H\times W, m=\underset{i}{\text{argmax}}(S_{i,j})\}.
\end{equation}

\subsubsection{Backbone}
Subsequently, the primary RGB data and the dynamically selected auxiliary modality are encoded by parallel Transformer backbones (MiT)~\cite{xie2021segformer}. The extracted multi-level features are then passed to the proposed stage-wise fusion modules. The process can be formulated as follows:
\begin{equation}
    x_\text{r}=\text{MiT}(x_{\text{r}}^{(0)}), \, x_\text{a}=\text{MiT}(x_{\text{a}}^{(0)}).
\end{equation}
\Needspace{230pt}
\subsection{Bidirectional Polarity-Aware Cross-Modality Fusion Module}

%\begin{figure}[t!]
%  \centering
%  \includegraphics[width=\linewidth]{figures/fusion.pdf}
%  \caption[Fusion Component]{Overview of our proposed fusion component. The Bidirectional Cross-Branch Interaction module first exchanges information across modalities through polarity-aware cross attention. The Normalization-Modulated Refinement (NMR) then performs normalization-based modulation and feature refinement with stage-wise alignment loss $\mathcal{L}_{\text{stage}}$, producing the final fused features $y_{\text{fused}}$.}
%  \label{fig:fusion}
%\end{figure}

% \begin{paracol}{2}

\begin{wrapfigure}{r}{0.44\textwidth}
    \centering

  \includegraphics[width=\linewidth]{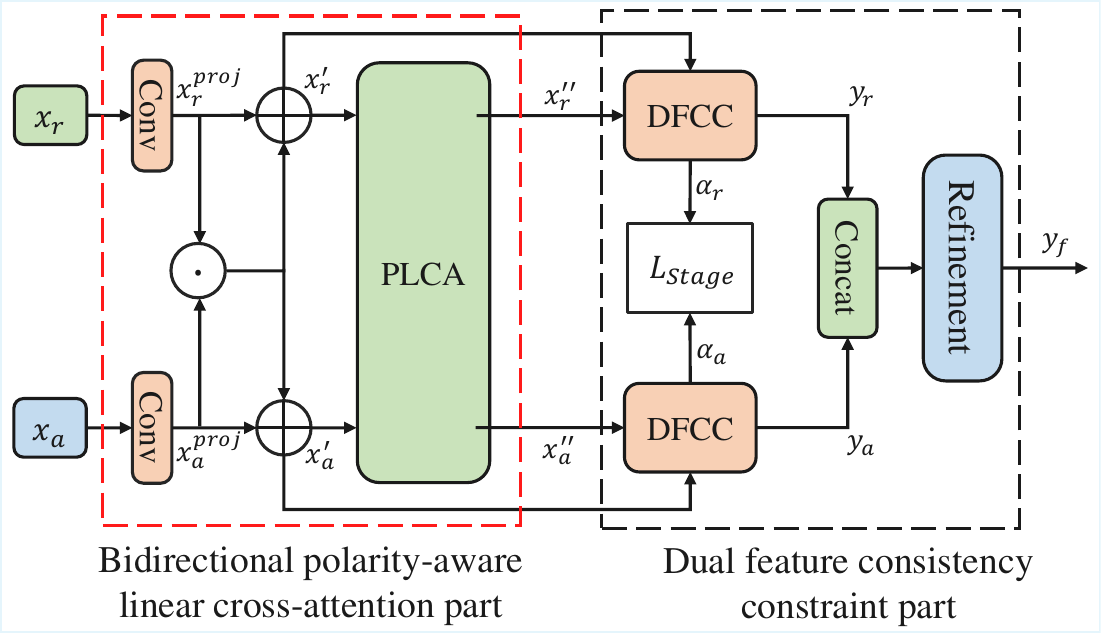}
  \caption[figure]{The overall structure of the BPCF module. It consists of a bidirectional polarity-aware linear cross-attention (BPLCA) part and a dual feature consistency constraint (DFCC) part.}
  \label{fig:fusion}
\end{wrapfigure}

    To effectively integrate heterogeneous cues from RGB and auxiliary modalities, we introduce a bidirectional polarity-aware cross-modality fusion (BPCF) module, as illustrated in Fig.~\ref{fig:fusion}. The BPCF is composed of two key parts: bidirectional polarity-aware linear cross-attention (BPLCA) and dual feature consistency constraint (DFCC). The BPLCA facilitates comprehensive cross-modal interaction by employing symmetric cross-gating and polarity-aware cross-linear attention. Meanwhile, the DFCC enhances feature alignment by exploiting statistical correlations and applying the consistency constraint loss $\mathcal{L}_{\text{stage}}$, complemented by a refinement block to generate the final fused output.
    
% \switchcolumn

% \end{paracol}

%To effectively integrate heterogeneous cues from RGB and auxiliary modalities, we design a two-stage fusion component as illustrated in Fig.~\ref{fig:fusion}. \textbf{Bidirectional Cross-Branch Interaction} enables expressive cross-modal exchange through symmetric cross-gating and polarity-aware attention. \textbf{Normalization-Modulated Refinement (NMR)} further aligns the fused features by leveraging normalization-based modulation and a refinement block, while enforcing stage-wise consistency via $\mathcal{L}_{\text{stage}}$. Together, these modules provide a unified framework that balances efficiency, semantic alignment, and representational richness.

 \subsubsection{Bidirectional polarity-aware linear cross-attention (BPLCA)} 
\begin{wrapfigure}{r}{0.40\textwidth}
    \centering

  \includegraphics[width=\linewidth]{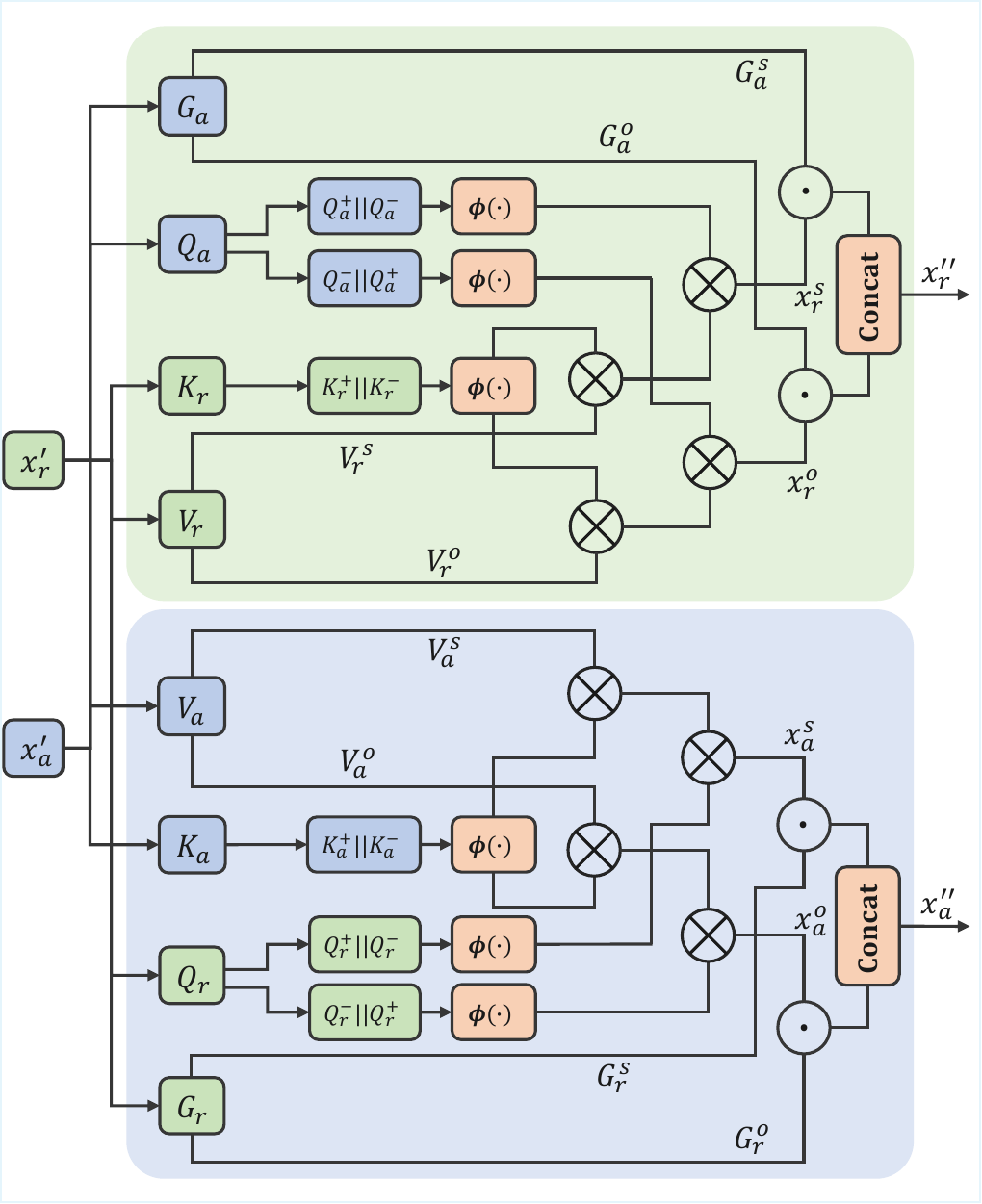}
  \caption{Illustration of BPLCA. Features from RGB and auxiliary modality are deeply fused through the BPLCA component. }
  \label{fig:plca}
\end{wrapfigure}

 To comprehensively fuse complementary cues from the RGB stream and the auxiliary modality stream, our BPLCA is designed with dual cross-modal fusion branches. First, an efficient cross-gating mechanism is applied to activate relevant features. Second, a symmetric polarity-aware linear cross-attention (PLCA) is introduced (see Fig.~\ref{fig:plca}), which preserves informative negative evidence while maintaining the linear-time and linear-memory efficiency of kernelized attention. Formally, let $x_{\mathrm{r}},x_{\mathrm{a}}\in\mathbb{R}^{N\times C}$ represent the extracted features from the RGB and auxiliary branches, reshaped into sequences of $N{=}HW$ tokens with $C$ channels. Two $1{\times}1$ convolutions are then used to perform per-location linear projections:
\wrapfinish

\begin{equation}
x_{\text{r}}^\text{proj}=\text{Conv}_{1\times1}(x_{\text{r}}), x_{\text{a}}^\text{proj}=\text{Conv}_{1\times1}(x_{\text{a}}).
\end{equation}
We then realize bidirectional interaction via symmetric cross-gating using the element-wise (Hadamard) product $\circ$:
\begin{equation}
    \begin{aligned}
x'_{\text{r}}=x_{\text{r}}^\text{proj} + (x_{\text{r}}^\text{proj}\circ x_{\text{a}}^\text{proj}),
\\
x'_{\text{a}}=x_{\text{a}}^\text{proj} + (x_{\text{r}}^\text{proj}\circ x_{\text{a}}^\text{proj}).
    \end{aligned}
\end{equation}
To further enhance cross-modal interaction, the fused features are passed into the PLCA, as illustrated in Fig.~\ref{fig:plca}. Within PLCA, the input features are projected into four distinct matrices: query ($Q$), key ($K$), value ($V$), and gating ($G$). Among these, the gating matrix $G$ plays a pivotal role in realizing polarity-aware attention~\cite{Meng2025PolaFormerPL}, whose mechanism will be detailed in the following section.
\begin{equation}
    \begin{aligned}
        Q_\text{r}=x'_\text{r}q_\text{r},\ 
        K_\text{r}=x'_\text{r}k_\text{r},\ 
        V_\text{r}=x'_\text{r}v_\text{r},\ 
        G_\text{r}=x'_\text{r}g_\text{r};\\
        Q_\text{a}=x'_\text{a}q_\text{a},\ 
        K_\text{a}=x'_\text{a}k_\text{a},\ 
        V_\text{a}=x'_\text{a}v_\text{a},\ 
        G_\text{a}=x'_\text{a}g_\text{a}. 
    \end{aligned}
\end{equation}

% Here, the cross-term $x_{\text{r}}^\text{proj}\circ x_{\text{a}}^\text{proj}$ reweights each pixel–channel response of one branch by the co-located response of the other, enabling early spatially aligned exchange without the overhead of explicit cross-attention. This operation is permutation-free over spatial positions, adds negligible parameters, and preserves the original activations through a residual path, which stabilizes optimization.

% \begin{paracol}{2}

where $q_\text{r}, k_\text{r}, v_\text{r}, g_\text{r}, q_\text{a}, k_\text{a}, v_\text{a}, g_\text{a}$ denote the learnable linear projections for the RGB and auxiliary branches, respectively. 
Following the kernelization strategy in~\cite{kat2020linearatt}, vanilla attention can be reformulated using a similarity function defined by a feature map $\phi(\cdot)$. The $i$-th row of the attention result is formulated as:
\begin{equation}\label{eq:genatt}
\text{Attention}(Q,K,V)_i=\frac{\sum_j \text{sim}(Q_i, K_j)V_j}{\sum_j \text{sim}(Q_i, K_j)},
\end{equation}
\begin{equation}\label{eq:ker}
\text{sim}(x, y) = \kappa(x, y) = \phi(x)^T\phi(y).
\end{equation}
This kernelization renders the similarity function decomposable. By exploiting distributive and associative properties, the final attention computation becomes:
\begin{equation}\label{eq:linatt}
    \text{Attention}(Q, K, V) = \frac{\phi(Q)^T\Bigl(\sum_j \phi(K_j)V_j\Bigr)}{\phi(Q)^T\Bigl(\sum_j \phi(K_j)\Bigr)}.
\end{equation}

% \end{paracol}

Notably, Eq.~\ref{eq:linatt} avoids the $\mathcal{O}(N^2D)$ pairwise computations $Q_i^\top K_j$ by precomputing the terms $\sum_j \phi(K_j)V_j$ and $\sum_j \phi(K_j)$, thereby reducing the overall time complexity to $\mathcal{O}(ND^2)$.
Standard non-negative feature maps like ReLU discard negative information. This information is retained by decomposing queries and keys into positive ($Q^+$) and negative ($Q^-$) parts via $Q=Q^+-Q^-$ where $Q^+=\text{ReLU}(Q)$ and $Q^-=\text{ReLU}(-Q)$. 
The same decomposition is applied to the key matrix $K$. Then the inner product can be decomposed into: 
\begin{equation} \label{eq:innerdecom}
\begin{aligned}
\langle q_i,k_j\rangle = &\langle q_i^+,k_j^+\rangle +\langle q_i^-,k_j^-\rangle \\
                        \ \ &-\langle q_i^-,k_j^+\rangle -\langle q_i^+,k_j^-\rangle. 
\end{aligned}
\end{equation}
Then the linear-decomposable similarity function in Eq.~\ref{eq:ker} can be derived as 
\begin{equation}\label{eq:simdec}
\begin{aligned}
    \text{sim}(x, y)=&\phi(q_i)^T\phi(k_j) \\
                    =&((\phi(q_i)^+)^T\phi(k_j)^+\\
                    &\ \ \ \ +(\phi(q_i)^-)^T\phi(k_j)^-) \\
                    &\ -((\phi(q_i)^+)^T\phi(k_j)^- \\
                    &\ \ \ \ \ \ \ \  +(\phi(q_i)^-)^T\phi(k_j)^+). 
\end{aligned}
\end{equation}
Thus, we can accept a non-positive feature map $\phi$ in the kernel. To process these non-negative components, we employ a channel-wise learnable feature map, $\phi(x)$, defined as:
\begin{equation}
\phi(x)=x^P, \text{where } P = 1+\alpha\cdot\text{sigmoid}(W).
\end{equation}
where $P$ is a dynamically computed exponent, determined by a hyperparameter $\alpha$ and a $d$-dimensional learnable parameter vector $W$, which learns to weigh the relative importance of different channels.
\wrapfinish

To explicitly model the interactions between polarities, we first concatenate the positive and negative components along the channel dimension (denoted by $\mathbin\Vert$), then capture the interactions between $Q$ and $K$ for \textit{same-signed} ($++$, $--$) and \textit{opposite-signed} ($+-$, $-+$) pairs separately while ensuring all kernelized terms remain non-negative. The value ($V$) is partitioned accordingly with $V_\text{r} = V_\text{r}^\text{s}\mathbin\Vert V_\text{r}^\text{o}$ and $V_\text{a} = V_\text{a}^\text{s}\mathbin\Vert V_\text{a}^\text{o}$, while both split tensors have half of the channels from the original one. Based on these decompositions, we then compute the intermediate fusion results for the RGB and auxiliary modalities, distinguishing between \textit{same-signed} and \textit{opposite-signed} correlations, substituting Eq.~\ref{eq:simdec} into Eq.~\ref{eq:linatt}:
\begin{equation}\label{eq:bibranch}
\begin{aligned}
    x_\text{r}^\text{s}&=\frac{\phi(Q_\text{a}^+\mathbin\Vert Q_\text{a}^-)^T\sum_j\phi(K_\text{r}^+\mathbin\Vert K_\text{r}^-)_j(V_{\text{r}}^\text{s})_j}{\phi(Q_\text{a}^+\mathbin\Vert Q_\text{a}^-)^T\sum_j\phi(K_\text{r}^+\mathbin\Vert K_\text{r}^-)_j},\\
    x_\text{r}^\text{o}&=\frac{\phi(Q_\text{a}^-\mathbin\Vert Q_\text{a}^+)^T\sum_j\phi(K_\text{r}^+\mathbin\Vert K_\text{r}^-)_j(V_{\text{r}}^\text{o})_j}{\phi(Q_\text{a}^-\mathbin\Vert Q_\text{a}^+)^T\sum_j\phi(K_\text{r}^+\mathbin\Vert K_\text{r}^-)_j};\\
    x_\text{a}^\text{s}&=\frac{\phi(Q_\text{r}^+\mathbin\Vert Q_\text{r}^-)^T\sum_j\phi(K_\text{a}^+\mathbin\Vert K_\text{a}^-)_j(V_{\text{a}}^\text{s})_j}{\phi(Q_\text{r}^+\mathbin\Vert Q_\text{r}^-)^T\sum_j\phi(K_\text{a}^+\mathbin\Vert K_\text{a}^-)_j},\\
    x_\text{a}^\text{o}&=\frac{\phi(Q_\text{r}^-\mathbin\Vert Q_\text{r}^+)^T\sum_j\phi(K_\text{a}^+\mathbin\Vert K_\text{a}^-)_j(V_{\text{a}}^\text{o})_j}{\phi(Q_\text{r}^-\mathbin\Vert Q_\text{r}^+)^T\sum_j\phi(K_\text{a}^+\mathbin\Vert K_\text{a}^-)_j}.\\
\end{aligned}
\end{equation}
To capture sophisticated relations between \textit{same-signed} and \textit{opposite-signed} parts, we employ a learnable, element-wise mixing that \emph{weights} the \textit{same-signed} and \textit{opposite-signed} streams via gating tensors, $G=G^\text{s}\mathbin\Vert G^\text{o}$. The final polarity-aware outputs are then produced through gated fusion:
\begin{equation}
\begin{aligned}
x''_\text{r}=(x_\text{r}^\text{s}\circ G_\text{a}^\text{s})\mathbin\Vert (x_\text{r}^\text{o}\circ G_\text{a}^\text{o}),
\\
x''_\text{a}=(x_\text{a}^\text{s}\circ G_\text{r}^\text{s})\mathbin\Vert (x_\text{a}^\text{o}\circ G_\text{r}^\text{o}).
\end{aligned}
\end{equation}
% Notably, the query matrix $Q$ and gating matrix $G$ are designed to interact with the other modality in the \textit{cross-modality} design of the PLCA component. 
Other than the self attention for enhanced \textit{uni-modal image perception} originated from PolaFormer~\cite{Meng2025PolaFormerPL}, we emphasize the high-sensitivity \textit{cross-modal information aggregation} of our polarity-aware cross attention. 
Especially, PLCA incorporates a bidirectional architecture where the Query ($Q$) and Gating ($G$) matrices of one branch are explicitly designed to interact with the Key ($K$) and Value ($V$) of the other modality, both achieving linear complexity and capturing universal signals. 
This mechanism preserves the integrity of the original content alignment by mitigating excessive influence from the complementary modality, yet still allows for the infusion of cross-modal information. 
% \subsubsection{Normalization-Modulated Refinement (NMR)}
% \label{sec:nmr}

\Needspace{235pt}
\subsubsection{Dual feature consistency constraint (DFCC)}
\begin{wrapfigure}{r}{0.44\textwidth}
    \centering
    \vspace{-1em}
  \includegraphics[width=\linewidth]{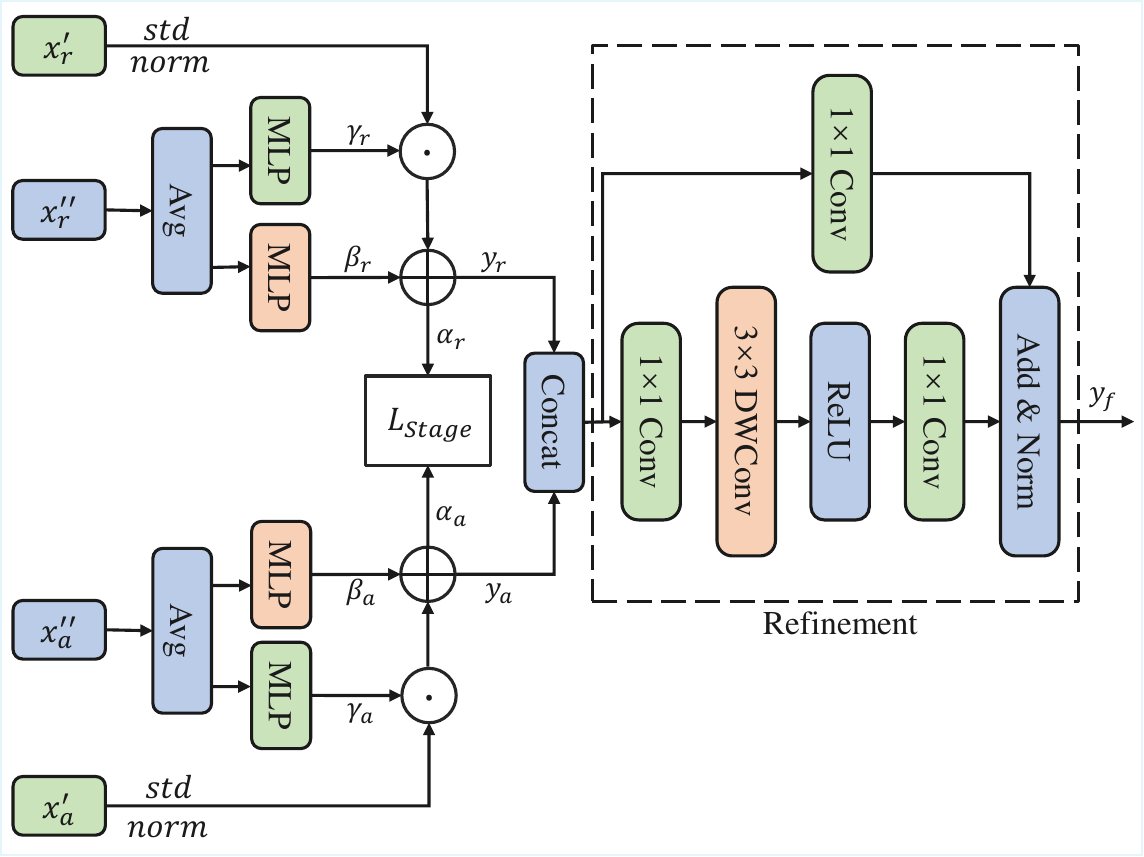}
  \caption{Illustration of DFCC. The features from BPLCA are further refined to obtain the final fused output.}
  \label{fig:nmr}
\end{wrapfigure}

 To strengthen cross-modal feature alignment while maintaining consistency, we introduce the DFCC. As shown in Fig.~\ref{fig:nmr}, DFCC integrates a feature consistency constraint with a refinement block in a unified design, jointly enabling semantic alignment and feature enhancement. Specifically, given enhanced RGB and auxiliary features $x_{\text{r}}',\; x_{\text{a}}' \in \mathbb{R}^{B \times C \times H \times W}$ and fused features $x_{\text{r}}'',\; x_{\text{a}}'' \in \mathbb{R}^{B \times C \times H \times W}$, we first compute feature statistics to align the feature spaces across both streams:
\begin{equation}
    \begin{aligned}
    \hat{x}_{\text{r}} =
    \frac{x_{\text{r}}' - \mu(x_{\text{r}}')}
         {\sigma(x_{\text{r}}')}, 
    \\
    \hat{x}_{\text{a}} =
    \frac{x_{\text{a}}' - \mu(x_{\text{a}}')}
         {\sigma(x_{\text{a}}')};
    \end{aligned}
\end{equation}

\vspace{0.5em}
where, 

\wrapfinish
% \vspace{1em}
\begin{equation}
    \begin{aligned}
    \mu(x)= \frac{1}{HW} \sum_{h=1}^{H}\sum_{w=1}^{W} x,\ \ 
\sigma(x) = \sqrt{\frac{1}{HW} \sum_{h=1}^{H}\sum_{w=1}^{W} \big(x - \mu(x)\big)^2 }.
    \end{aligned}
\end{equation}
where $\mu(x)$ and $\sigma(x)$ denote the channel-wise mean and standard deviation, respectively.

DFCC then learns adaptive affine modulation parameters from the fused features $x_{\text{r}}''$ and $x_{\text{a}}''$. Specifically, the global average pooling operations are first applied:
\begin{equation}
    z_{\text{r}} = \text{AvgPool}(x_{\text{r}}''), 
    z_{\text{a}} = \text{AvgPool}(x_{\text{a}}'') .
\end{equation}
Subsequently, MLPs are employed to generate branch-specific modulation parameters:
\begin{equation}
\begin{aligned}
    \gamma_{\text{r}}= \text{MLP}(z_{\text{r}}), 
    \beta_{\text{r}}  = \text{MLP}(z_{\text{r}}),
    \\
\gamma_{\text{a}}= \text{MLP}(z_{\text{a}}), 
    \beta_{\text{a}}  = \text{MLP}(z_{\text{a}}).
\end{aligned}
\end{equation}
The normalized features are then modulated as:
\begin{equation}
    y_{\text{r}} = \gamma_{\text{r}} \cdot \hat{x}_{\text{r}} + \beta_{\text{r}}, 
    y_{\text{a}} = \gamma_{\text{a}} \cdot \hat{x}_{\text{a}} + \beta_{\text{a}}.
\end{equation}
Finally, the two streams are concatenated and fed into the refinement block, which comprises two parallel paths: a $1 \times 1$ convolutional shortcut that preserves identity information, and a nonlinear transformation branch that enhances feature fusion. The nonlinear branch sequentially applies a $1 \times 1$ convolution for channel reduction, a $3 \times 3$ depthwise convolution for spatial interaction, a ReLU activation, and another $1 \times 1$ convolution to restore the channel dimension, followed by normalization. The final output is then obtained by summing the two paths: 
\begin{equation}
    y_{\text{f}} = \text{Refinement}(y_{\text{r}} \mathbin\Vert y_{\text{a}}).
\end{equation}

% where,
% \[
%     \text{Refinement}(u) = \text{Norm}\left(\text{Conv}_{1 \times 1}(y_{\text{r}} \mathbin\Vert y_{\text{a}}) 
%     + \text{Conv}_{1 \times 1}\big(\text{ReLU}(\text{DWConv}_{3 \times 3}(\text{Conv}_{1 \times 1}(y_{\text{r}} \mathbin\Vert y_{\text{a}})))\big)\right).
% \]

%\begin{figure}
%  \centering
%  \includegraphics[width=0.7\linewidth]{figures/NMR.pdf}
%  \caption[NMR]{Illustration of NMR. Feature-conditioned normalization provides modulation parameters, while the Refinement module integrates identity and nonlinear transformations for cross-modal alignment.}
%  \label{fig:nmr}
%\end{figure}

%where $\text{Norm}(a) = \frac{a - \min(a)}{\max(a) - \min(a) + \epsilon}$ is applied pixel-wise. The stage-wise alignment loss is then formulated as the mean squared error (MSE) between the normalized attention maps:
%\begin{equation}
%    \mathcal{L}_{\text{stage}} = \| \alpha_{\text{r}} - \alpha_{\text{a}} \|_2^2.
%\end{equation}
%This objective encourages both modalities to produce consistent attention responses, thereby improving feature alignment within the Refinement process.

%\vspace{0.5em}
%\noindent
%Overall, NMR provides a principled normalization-modulated refinement mechanism that ensures consistent cross-modal feature integration. This unified design complements our fusion module by supplying semantically aligned, spatially coherent features for downstream decoding.

Moreover, to further enhance cross-modal consistency, we incorporate a stage-wise constraint loss within DFCC. At each fusion stage, the features $y_{\text{r}}$ and $y_{\text{a}}$ are computed by averaging over the channel dimension:
\begin{equation}
\begin{aligned}
    \alpha_{\text{r}} &= \text{Norm}(\text{Mean}(y_{\text{r}}, \text{dim}=1)),\\
    \alpha_{\text{a}} &= \text{Norm}(\text{Mean}(y_{\text{a}}, \text{dim}=1)).
\end{aligned}
\end{equation}
The $\text{Norm}(a) = \frac{a - \min(a)}{\max(a) - \min(a) + \epsilon}$ here is applied pixel-wise. The stage-wise constraint loss is then defined as the mean squared error (MSE) between the normalized features:
\begin{equation}
    \mathcal{L}_{\text{stage}} = \| \alpha_{\text{r}} - \alpha_{\text{a}} \|_2^2.
\end{equation}
This objective encourages both modalities to produce consistent responses, thereby improving feature alignment.
\subsection{Loss Function}
\label{sec:loss_function}

Our proposed model is trained using a combination of segmentation and alignment objectives. For the primary segmentation task, we employ the standard pixel-wise cross-entropy loss, commonly used in dense prediction tasks:
\begin{equation}
\mathcal{L}_{\text{CE}} = - \frac{1}{N} \sum_{i=1}^{N} \log \frac{\exp(p_{i, y_i})}{\sum_{j=1}^{C} \exp(p_{i, j})}.
\end{equation}
where $N = H \times W$ denotes the number of valid pixels, $C$ is the number of classes, $p_{i,j}$ represents the logit for class $j$ at pixel $i$, and $y_i$ is the corresponding ground-truth label.
The overall training objective combines the segmentation loss with the stage-wise constraint loss as follows:
\begin{equation}
\mathcal{L}_{\text{total}} = \lambda_1 \cdot \mathcal{L}_{\text{CE}} + \lambda_2 \cdot \sum_{i=1}^{4} \mathcal{L}_{\text{stage}}^{(i)}.
\end{equation}
where $\lambda_1$ and $\lambda_2$ balance the segmentation and stage-wise constraint losses. In our experiments, we set $\lambda_1 = 1$ and $\lambda_2 = 0.1$.

%% file: sections/04_experiments.tex
% \section{Results}

\section{Experiments}

% To validate the effectiveness of the proposed method, we conduct extensive experiments across eight multi-modal datasets. The detailed analysis is presented below.
\subsection{Experimental Setup}
\label{sec:experimental_setup}
\subsubsection{Datasets}
\label{sec:datasets}
\label{app:datasetdetails}
We evaluate our approach on eight widely used multi-modal datasets: MFNet~\cite{ha2017mfnet}, KITTI-360~\cite{liao2022kitti}, DDD17~\cite{Binas2017DDD17ED}, DSEC~\cite{gehrig2021dsec}, DELIVER~\cite{zhang2023delivering}, UrbanLF~\cite{sheng2022urbanlf}, MCubeS~\cite{liang2022multimodal}, and ZJU~\cite{xiang2021polarization}. An overview of the datasets is provided in Tab.~\ref{tab:dataset_overview}; additional dataset details and common configuration parameters are provided in Sec.~\ref{app:datasetdetails} and Sec.~\ref{app:impdetails}.
\begin{table*}[tbp]
\centering
\compacttablefont
\caption[Datasets overview]{Summary of multi-modal datasets used in our experiments.}
\fitbox{
\begin{tabular}{lcccc}
\toprule
Dataset & Modality & Classes & Train / Test& Resolution \\
\midrule
MFNet~\cite{ha2017mfnet}        & RGB-Thermal   & 8   & 784 / 392   & $480\times640$  \\
KITTI-360~\cite{liao2022kitti}   & RGB-LiDAR     & 19   & 49004 / 12276& $1408\times376$  \\
DELIVER~\cite{zhang2023delivering}   & RGB-Depth-Event-LiDAR   & 25   & 3893 / 1897 & $1024\times1024$   \\
DDD17~\cite{Binas2017DDD17ED}   & RGB-Event     & 5   & 15950 / 3890& $200\times346$  \\
DSEC~\cite{gehrig2021dsec}      & RGB-Event     & 11  & 8082 / 2809 & $440\times640$  \\
MCubeS~\cite{liang2022multimodal} & RGB-AoLP-DoLP-NIR & 20 & 302 / 102   & $1224\times1024$ \\
ZJU~\cite{xiang2021polarization} & RGB-Polarization & 8 & 344 / 50   & $1224\times1024$ \\
UrbanLF~\cite{sheng2022urbanlf} & RGB-Light Field & 14 & 580 / 80 (Real), 172 / 28 (Syn) & $623\times432$, $640\times480$ \\
\bottomrule
\end{tabular}
}
\label{tab:dataset_overview}
\end{table*}

An overview of the datasets used in our experiments is provided in Tab.~\ref{tab:dataset_overview}. The details are introduced below:

\textbf{RGB-T MFNet}~\cite{ha2017mfnet} is a driving-scene dataset with 1,569 aligned RGB-thermal pairs across 8 semantic classes. It provides 784/392/392 images for training, validation, and testing, respectively, covering both daytime and nighttime conditions.  

\textbf{RGB-L KITTI-360}~\cite{liao2022kitti} is a suburban
driving dataset, providing 49,004 training and 12,276 validation images with a resolution of $1408 \times 376$. Following the Cityscapes dataset~\cite{cordts2016cityscapes}, it defines 19 semantic classes.

\textbf{RGB-X DELIVER}~\cite{zhang2023delivering} is a large-scale synthetic benchmark built in the CARLA simulator, supporting multi-modal segmentation across RGB, depth, event, and LiDAR streams. It contains over 47k annotated frames from six camera views under diverse conditions, including challenging weather (fog, rain, night) and five sensor degradation types (e.g., motion blur, exposure imbalance, LiDAR jitter). The dataset defines 25 semantic classes covering urban elements such as vehicles, roads, pedestrians, and vegetation. 

\textbf{RGB-E DDD17}~\cite{Binas2017DDD17ED} includes over 12 hours of driving data collected with a DAVIS346B sensor, providing event streams and grayscale images at $200 \times 346$ resolution. We generate pseudo-labels using EV-SegNet~\cite{alonso2019ev}, resulting in 5 semantic classes and splits of 15,950/3,890 samples.  

\textbf{RGB-E DSEC}~\cite{gehrig2021dsec} contains over 10k RGB-event frames with 11 categories, captured from stereo cameras in urban and rural environments. We follow~\cite{xie2024eisnet} for preprocessing and dataset splits (8082/2809 for training/testing).  

\textbf{RGB-P MCubeS}~\cite{liang2022multimodal} consists of 500 samples with aligned RGB, NIR, and polarization cues (DoLP, AoLP), annotated with 20 material categories. The dataset is split into 302/96/102 for training, validation, and testing.

\textbf{RGB-P ZJU}~\cite{xiang2021polarization} is a polarization semantic segmentation dataset collected in outdoor scenes. 
Each sample contains four polarized RGB images at $0^\circ$, $45^\circ$, $90^\circ$, and $135^\circ$, from which AoLP and DoLP are derived using Stokes parameters. It provides 344 training and 50 validation samples across 8 semantic classes, with image resolution of $1224 \times 1024$.

\textbf{RGB-LF UrbanLF}~\cite{sheng2022urbanlf} is a light-field dataset for urban-scene segmentation, including both real (580/80/164) and synthetic (172/28/50) subsets, each with 14 semantic classes. Central-view annotations are used for supervision.

\wrapfinish

\subsubsection{Implementation Details}
\label{sec:implementation}

\begin{table*}[tbp]
\centering
\compacttablefont
\caption[Dataset specific configuration parameters]{Dataset-specific training configuration parameters.}
\renewcommand{\arraystretch}{1.05}
\fitbox{
\begin{tabular}{lccccc}
\toprule
\textbf{Dataset} & \textbf{Shrink} & \textbf{Shrink Scale} & \textbf{Backbone} & \textbf{Learning Rate} & \textbf{Training Epochs} \\
\midrule
MFNet~\cite{ha2017mfnet}               & True  & 0.9 & mit\_b4 & 6e-5 & 500 \\
KITTI-360~\cite{liao2022kitti}   & False     & -   & mit\_b2 & 6e-5 & 60  \\
DELIVER~\cite{zhang2023delivering} & True  & 0.5 & mit\_b2 & 6e-5 & 60 \\
DDD17~\cite{Binas2017DDD17ED} & False & -   & mit\_b4 & 2e-4 & 40 \\
DSEC~\cite{gehrig2021dsec}             & False & -   & mit\_b4 & 6e-5 & 60 \\
MCubeS~\cite{liang2022multimodal}      & False & -   & mit\_b4 & 6e-5 & 500 \\
ZJU~\cite{xiang2021polarization}       & False & -   & mit\_b4 & 6e-5 & 500 \\
UrbanLF~\cite{sheng2022urbanlf}        & False & -   & mit\_b4 & 6e-5 & 500 \\
\bottomrule
\end{tabular}
}

\label{tab:dataset_config}
  
\end{table*}

\label{app:impdetails}
\begin{wraptable}{r}{0.44\textwidth}
\centering
\compacttablefont
\caption[Common configuration parameters]{Common configuration hyperparameters.}
\fitbox{
\begin{tabular}{ll}
\toprule
\textbf{Parameter} & \textbf{Value} \\
\midrule
Background & 255 \\
Learning Rate (LR) Power & 0.9 \\
Momentum & 0.9 \\
Weight Decay & 0.01 \\
Batch Size & 2 \\
Number of Workers & 2 \\
Train Scale Array & {[0.5, 0.75, 1, 1.25, 1.5, 1.75]} \\
Warm-up Epochs & 10 \\
Fix Bias & True \\
Batch Norm Epsilon & 1e-3 \\
Batch Norm Momentum & 0.1 \\
Evaluation Iteration & 25 \\
Evaluation Stride Rate & 2/3 \\
Evaluation Scale Array & {[1]} \\
Evaluation Flip & False \\
\bottomrule
\end{tabular}
}
\label{tab:common_config}
\end{wraptable}

Except for the high-resolution DELIVER experiments in Sec.~\ref{sec:resolution_analysis}, models are trained on a single NVIDIA RTX 3090 GPU with a batch size of 2. We adopt AdamW~\cite{loshchilov2019decoupled} as the optimizer, with an initial learning rate ranging from $6 \times 10^{-5}$ to $2 \times 10^{-4}$ depending on the dataset, momentum of $0.9$, weight decay of $0.01$, and epsilon of $1 \times 10^{-8}$. The learning rate is scheduled using polynomial decay with a power of $0.9$, preceded by a 10-epoch linear warm-up. Input images are resized to dataset-specific resolutions, and standard data augmentations are applied, including random resizing within [0.5, 2.0], horizontal flipping, color jitter, Gaussian blur, and random cropping. Following prior works such as CMX~\cite{zhang2023cmx} and CMNeXt~\cite{zhang2023delivering}, we adopt dataset-specific training configurations to ensure fair comparisons, where backbone variants (e.g., MiT-B2 or MiT-B4), learning rates, and training epochs are selected according to commonly used settings for each dataset. This practice is widely adopted in RGB-X semantic segmentation due to differences in dataset scale, resolution, and modality characteristics. More details can be found in Tab.~\ref{tab:dataset_config} and Tab.~\ref{tab:common_config}.
\wrapfinish

\subsubsection{Evaluation Metrics}
\label{sec:evaluation_metrics}
\label{sec:metrics}
We evaluate model performance using a set of standard metrics for semantic segmentation, including per-class Intersection over Union (IoU), mean IoU (mIoU), mean IoU excluding the background class. These metrics jointly provide a comprehensive assessment of segmentation quality across categories and pixels, while also accounting for class imbalance. In addition, we report computational efficiency in terms of FLOPs, number of parameters, and inference speed (FPS) to evaluate the practicality of our models for real-world deployment.
Detailed metric definitions are provided below.

\paragraph{Intersection over Union (IoU):} IoU quantifies the overlap between the predicted and ground truth segmentation masks for each class. It is defined as:
 \begin{equation}
 \text{IoU}_c = \frac{|A_c \cap B_c|}{|A_c \cup B_c|},
 \end{equation}
 where $A_c$ and $B_c$ denote the predicted and ground truth pixels for class $c$, respectively. A higher IoU indicates better boundary delineation and region matching.

 \paragraph{Mean IoU (mIoU):} The primary evaluation metric is mIoU, which averages IoU scores across all classes, ensuring a balanced evaluation regardless of class frequency. Formally,
 \begin{equation}
 \text{mIoU} = \frac{1}{C} \sum_{c=1}^{C} \frac{|A_c \cap B_c|}{|A_c \cup B_c|},
 \end{equation}
 where $C$ is the total number of classes. This metric is particularly valuable in datasets with diverse categories, such as autonomous driving, where rare classes (e.g., pedestrians) must be treated with equal importance as dominant ones (e.g., road).

% \paragraph{Frequency Weighted IoU (FreqIoU):} FreqIoU incorporates class frequency into the IoU computation, emphasizing the performance on common classes. It is defined as:
% \begin{equation}
% \text{FWIoU} = \sum_{c=1}^{C} \frac{n_c}{N} \cdot \frac{|A_c \cap B_c|}{|A_c \cup B_c|},
% \end{equation}
% where $n_c$ is the number of ground truth pixels for class $c$ and $N$ is the total number of pixels. This metric highlights model performance under class imbalance, which is typical in road-scene datasets.

% \paragraph{Pixel Accuracy (PA):} Pixel accuracy reflects the ratio of correctly classified pixels over the entire dataset:
% \begin{equation}
% \text{PA} = \frac{\text{Correctly Classified Pixels}}{\text{Total Number of Pixels}}.
% \end{equation}
% While intuitive, this metric may overestimate performance if dominant classes occupy large portions of the image.

% \paragraph{Mean Pixel Accuracy (mPA):} mPA evaluates pixel-level classification accuracy within each class, averaged across classes:
% {\fontsize{9bp}{11bp}\selectfont
% \begin{equation}
% \text{mPA} = \frac{1}{C} \sum_{c=1}^{C} \frac{\text{Correctly Classified Pixels in Class } c}{\text{Total Pixels in Class } c}.
% \end{equation}
% }

% This ensures that minority classes are not overshadowed by dominant ones, complementing PA and mIoU for a fairer evaluation.

 \paragraph{Computational Efficiency Metrics:}To assess efficiency, we additionally report the computational cost of each model, including floating-point operations (FLOPs), number of parameters, and inference speed measured in frames per second (FPS). FLOPs estimate the theoretical workload of a forward pass, parameter count reflects model size and memory footprint, and FPS measures practical deployment speed under fixed input resolution.
 By jointly analyzing these metrics, we achieve a comprehensive evaluation of both segmentation quality and computational efficiency.

\Needspace{240pt}
\subsection{Quantitative results}

\Needspace{220pt}
\subsubsection{Results on MFNet}
\begin{wraptable}{r}{0.48\textwidth}
\centering
\compacttablefont
        \caption{Semantic segmentation results on the MFNet RGB-Thermal dataset.}
        \label{tab:mfnet_results}
\fitbox{%
\begin{tabular}{l|c|c|c}
                    \toprule
            \textbf{Method} & \textbf{Venue} & \textbf{Backbone} & \textbf{mIoU (\%)}  \\
            \midrule
            PAP~\cite{zhang2019pattern}& CVPR'19 & ResNet-18& 50.5 \\
            ABMDRNet~\cite{zhang2021abmdrnet} & CVPR'21 & ResNet-18 & 54.8  \\
            GMNet~\cite{zhou2021gmnet} & TIP'21 & ResNet-50 & 57.3  \\
            EGFNet~\cite{zhou2022edge} & AAAI'22 & ResNet-152 & 54.8  \\
            DooDLeNet~\cite{frigo2022doodlenet} & CVPR'22 & ResNet-101 & 57.3  \\
            CMX~\cite{zhang2023cmx} & TITS'23 & MiT-B4 & 59.7 \\
            CMNeXt~\cite{zhang2023delivering} & CVPR'23 & MiT-B4 & 59.9 \\
            CAFR~\cite{cao2024embracing}  & ECCV'24 & MiT-B4 & 67.2\\
            KTBNet~\cite{cai2025keep} & CVPR'25 & Swin-B& 59.9 \\
            OmniSegmentor~\cite{yin2025omnisegmentor} & NeurIPS'25& DFormer-L& 60.6\\
            \midrule
            Ours & - & MiT-B4 & \textbf{76.9} \\
            \bottomrule
            \end{tabular}%
}
\end{wraptable}

We evaluate our method on the MFNet dataset, a popular benchmark for RGB-thermal semantic segmentation. As shown in Tab.~\ref{tab:mfnet_results}, our model achieves 76.9\% mIoU, significantly outperforming prior methods such as OmniSegmentor~\cite{yin2025omnisegmentor} (60.6\%), KTBNet~\cite{cai2025keep} (59.9\%), and CAFR~\cite{cao2024embracing} (67.2\%). In particular, our approach yields a \textbf{9.7\%} absolute improvement over CAFR, highlighting its superior capability for RGB-thermal segmentation. We attribute this strong gain to the fact that RGB-thermal fusion often contains both shared semantics and complementary / inverse responses across modalities, especially under adverse illumination where RGB cues degrade but thermal remains informative. Our BPLCA explicitly models both same-signed and opposite-signed cross-modal correlations by decomposing the query/key features into positive and negative components and fusing them in a bidirectional manner, which allows the network to preserve informative ``negative evidence'' rather than discarding it via non-negative attention mappings. In addition, DFCC and the stage-wise constraint loss $L_{\text{stage}}$ promote cross-modal feature alignment and stabilize stage-wise fusion, alleviating sensor heterogeneity and reducing the risk of over-relying on a single modality. These mechanisms are consistent with our ablation findings that the fusion and alignment components are complementary and lead to the best MFNet performance when combined.

\wrapfinish

\Needspace{150pt}
\subsubsection{Results on KITTI-360}
\begin{wraptable}{r}{0.48\textwidth}
\centering
\compacttablefont
        \caption{Semantic segmentation results on the KITTI-360 RGB-LiDAR dataset.}
        \label{tab:rgb_lidar_results}
\fitbox{%
\begin{tabular}{l|c|c|c}
                    \toprule
            \textbf{Method} & \textbf{Venue} & \textbf{Backbone}  & \textbf{mIoU (\%)} \\
            \midrule
            PMF~\cite{zhuang2021perception}& ICCV'21 & SalsaNext  & 54.5  \\
            TransFuser~\cite{Prakash2021MultiModalFT} & CVPR'21 & RegNetY & 56.6 \\
            TokenFusion~\cite{wang2022multimodal} & CVPR'22 & MiT-B2  & 54.6  \\
            HRFuser~\cite{broedermann2023hrfuser} & ITSC'23 & HRFormer-T & 48.7 \\
            CMX~\cite{zhang2023cmx}  & TITS'23 & MiT-B2  & 64.3 \\
            CMNeXt~\cite{zhang2023delivering} & CVPR'23 & MiT-B2 & 65.3 \\
            CAFR~\cite{cao2024embracing} & ECCV'24 & MiT-B2 & 64.5  \\
            DFormer~\cite{yin2024dformer} & ICLR'24 & DFormer-L & 66.3\\
            \midrule
            Ours    & - & MiT-B2 & \textbf{66.8}  \\
            \bottomrule
            \end{tabular}%
}
\end{wraptable}

As shown in Tab.~\ref{tab:rgb_lidar_results}, our model achieves 66.8\% mIoU, outperforming all competing fusion methods. In particular, PMF~\cite{zhuang2021perception}, TransFuser~\cite{Prakash2021MultiModalFT}, TokenFusion~\cite{wang2022multimodal}, and HRFuser~\cite{broedermann2023hrfuser} lag significantly behind. Moreover, our approach surpasses DFormer~\cite{yin2024dformer} (66.3\%), CAFR~\cite{cao2024embracing} (64.5\%), and CMNeXt~\cite{zhang2023delivering} (65.3\%), further demonstrating the effectiveness of the proposed fusion mechanism. These results highlight the superiority of our method for RGB-LiDAR segmentation, particularly in large-scale outdoor scenes. We attribute the improvement on RGB-LiDAR segmentation to two aspects of our design. First, LiDAR provides complementary geometric structure while RGB provides dense texture/appearance cues; their feature statistics and spatial characteristics differ substantially, which makes naive fusion prone to mismatch.
Our DFCC explicitly aligns feature distributions via normalization and learnable affine modulation, and further enforces stage-wise consistency through $L_{\text{stage}}$, which encourages the two streams to produce coherent semantic responses across stages. Second, our fusion method performs stage-by-stage fusion, enabling progressive interaction between hierarchical representations rather than relying on a single late fusion point. Together, these components yield more stable cross-modal integration and improve segmentation quality in large-scale outdoor scenes.

\wrapfinish

\begin{table*}[tbp]
\centering
\compacttablefont
\caption{Semantic segmentation results on the DELIVER dataset.}
\label{tab:deliver_event_results}
\fitbox{%
\begin{tabular}{l|c|c|c|c}
\toprule
\textbf{Method} & \textbf{Venue} & \textbf{Modal} & \textbf{Backbone} & \textbf{mIoU} \\
\midrule
HRFuser~\cite{broedermann2023hrfuser} & ITSC'23 & RGB+D & HRFormer-T & 51.9 \\
TokenFusion~\cite{wang2022multimodal} & CVPR'22 & RGB+D & MiT-B2 & 60.3 \\
CMX~\cite{zhang2023cmx} & TITS'23 & RGB+D & MiT-B2 & 62.7 \\
CMNeXt~\cite{zhang2023delivering} & CVPR'23 & RGB+D & MiT-B2 & 63.6 \\
MemorySAM~\cite{liao2025memorysammemorizemodalitiessemantics} & Arxiv'25 & RGB+D & Hiera-B & 63.5 \\
Ours & - & RGB+D & MiT-B2 & \textbf{64.2} \\
\midrule
HRFuser~\cite{broedermann2023hrfuser} & ITSC'23 & RGB+E & HRFormer-T & 42.2 \\
TokenFusion & CVPR'22 & RGB+E & MiT-B2 & 45.6 \\
CMX~\cite{zhang2023cmx} & TITS'23 & RGB+E & MiT-B2 & 56.5 \\
CMNeXt~\cite{zhang2023delivering} & CVPR'23 & RGB+E & MiT-B2 & 57.5 \\
Any2Seg~\cite{zheng2024learning} & ECCV'24 & RGB+E & MiT-B2 & 57.8 \\
KTBNet~\cite{cai2025keep} & CVPR'25 & RGB+E & Swin-B & 58.4 \\
OmniSegmentor~\cite{yin2025omnisegmentor} & NeurIPS'25 & RGB+E & MiT-B2 & 58.4 \\
Ours & - & RGB+E & MiT-B2 & \textbf{60.3} \\
\midrule

HRFuser~\cite{broedermann2023hrfuser} & ITSC'23 & RGB+L & HRFormer-T & 43.1 \\
TokenFusion~\cite{wang2022multimodal} & CVPR'22 & RGB+L & MiT-B2 & 53.0 \\
CMX~\cite{zhang2023cmx} & TITS'23 & RGB+L & MiT-B2 & 56.4 \\
CMNeXt~\cite{zhang2023delivering} & CVPR'23 & RGB+L & MiT-B2 & 58.0 \\
Any2Seg~\cite{zheng2024learning} & ECCV'24 & RGB+L & MiT-B2 & 57.9 \\
OmniSegmentor~\cite{yin2025omnisegmentor} & NeurIPS'25 & RGB+L & MiT-B2 & 59.0 \\
Ours & - & RGB+L & MiT-B2 & \textbf{60.7} \\
\midrule
HRFuser~\cite{broedermann2023hrfuser} & ITSC'23 & RGB+D+L & HRFormer-T & 52.7 \\
CMNeXt~\cite{zhang2023delivering} & CVPR'23 & RGB+D+L & MiT-B2 & 65.5 \\
Ours & - & RGB+D+L & MiT-B2 & \textbf{65.6} \\
\midrule
HRFuser~\cite{broedermann2023hrfuser} & ITSC'23 & RGB+D+E+L & HRFormer-T & 53.0 \\
CMNeXt~\cite{zhang2023delivering} & CVPR'23 & RGB+D+E+L & MiT-B2 & 66.3 \\
MemorySAM~\cite{liao2025memorysammemorizemodalitiessemantics} & Arxiv'25 & RGB+D+E+L & Hiera-B & 65.4 \\
Ours & - & RGB+D+E+L & MiT-B2 & \textbf{67.4} \\
\bottomrule
\end{tabular}%
}
\end{table*}

\subsubsection{Results on DELIVER}
We further evaluate our approach on the DELIVER dataset, which provides high-resolution RGB-event pairs for semantic segmentation. We trained our model for a total of 60 epochs. As shown in Tab.~\ref{tab:deliver_event_results}, our method achieves excellent performance compared to other competitive methods. Even at half the original resolution and fewer training epochs, it surpasses recent SOTA methods. This result indicates that our fusion strategy remains effective under memory-constrained settings and is robust to resolution changes, because the proposed BPCF is applied in a stage-wise manner on multi-level features rather than depending on extremely high-resolution feature maps at a single stage.

\subsubsection{Results on DDD17 and DSEC}  
Tab.~\ref{tab:ddd17_dsec_results} summarizes the results on the DDD17 and DSEC datasets. Our model achieves 77.85\% mIoU and 97.10\% Pixel Acc on DDD17 dataset and 74.77\% mIoU and 95.60\% Pixel Acc on DSEC dataset. Compared with the second-best RGB+Event fusion method, EISNet~\cite{xie2024eisnet}, our approach consistently outperforms it in both mIoU and Pixel Acc. In contrast, event-only models such as ESEG-L~\cite{zhao2025eseg}, KWYAF~\cite{li2025know}, and OpenESS~\cite{kong2024openess} achieve considerably lower scores, underscoring the importance of multi-modal fusion. These results validate the effectiveness of our method and set a new SOTA on the DDD17 and DSEC datasets. We attribute the improvements on event-based driving datasets to the physical and statistical complementarity between RGB and events: events encode intensity changes and often exhibit response patterns that are not strictly positively correlated with RGB appearance. Our BPLCA is designed precisely to capture both same-signed and opposite-signed cross-modal correlations via polarity decomposition, enabling the model to exploit informative signals that would otherwise be suppressed by standard non-negative attention mappings. Meanwhile, DFCC and the stage-wise consistency loss encourage the two modalities to maintain aligned semantic responses across stages, which helps mitigate heterogeneous feature distributions and stabilizes optimization. Overall, the combination of polarity-aware interaction (BPLCA) and alignment-oriented regularization (DFCC+$L_{\text{stage}}$) leads to more reliable RGB-event fusion and consistent gains on both DDD17 and DSEC.
% Removed navigation to the DDD17 per-class table disabled by the author.
% Supplementary MFNet results for the 60-epoch setting are provided in Tab.~\ref{tab:mfnet_classwise}.

% The results for the DELIVER dataset are provided in the \textbf{Appendix}.

\begin{table*}[t!]
\centering
\caption[Results on DDD17~\cite{Binas2017DDD17ED} and DSEC~\cite{gehrig2021dsec}]{Semantic segmentation results on the DDD17 and DSEC RGB-Event datasets.}
\fitbox{ 
\begin{tabular}{l|c|c|cc|cc}
\toprule
\multirow{2}{*}{\textbf{Method}} &
\multirow{2}{*}{\textbf{Venue}} &
\multirow{2}{*}{\textbf{Modal}} & 
\multicolumn{2}{c|}{\textbf{DDD17}} &
\multicolumn{2}{c}{\textbf{DSEC}} \\
\cmidrule(lr){4-5}\cmidrule(lr){6-7}
& & & \textbf{mIoU (\%)} & \textbf{Acc. (\%)} & \textbf{mIoU (\%)} & \textbf{Acc. (\%)} \\
\midrule

% ---------------- Event ----------------
Ev-SegNet~\cite{alonso2019ev} & CVPRW'19 & Event & 54.81 & 89.76 & 51.76 & 88.61 \\
E2VID~\cite{rebecq2019events} & TRAMI'19 & Event & 48.47 & 85.84 & 44.08 & 80.06 \\
SegFormer~\cite{xie2021segformer} & NeurIPS'20 & Event & 52.76 & 89.34 & 54.19 & 90.12 \\
Vid2E~\cite{gehrig2020video} & CVPR'20 & Event & 56.01 & 90.19 & - & - \\
EvDistill~\cite{wang2021evdistill} & CVPR'21 & Event & 58.02 & - & - & - \\
DTL~\cite{wang2021dual} & ICCV'21 & Event & 58.80 & - & - & - \\
ESS~\cite{sun2022ess} & ECCV'22 & Event & 61.37 & 91.08 & 51.57 & 89.25 \\
EvSegFormer~\cite{jia2023event} & TIP'23 & Event & 54.41 & 94.72 & 56.04 & 90.38 \\
OpenESS~\cite{kong2024openess} & CVPR'24 & Event & 63.00 & 91.05 & 57.21 & 90.21 \\
KWYAF~\cite{li2025know} & AAAI'25 & Event & 57.69 & 90.04 & 57.75 & 90.87 \\
ESEG-B~\cite{zhao2025eseg} & AAAI'25 & Event & 57.01 & 89.64 & 55.93 & 90.22 \\
ESEG-L~\cite{zhao2025eseg} & AAAI'25 & Event & 59.97 & 90.68 & 57.55 & 91.47 \\
\midrule
% ---------------- RGB+Event ----------------
EDCNet-S2D~\cite{zhang2021exploring} & TITS'21 & RGB+Event & 61.99 & 93.80 & 56.75 & 92.39 \\
EV-Transfer~\cite{messikommer2022bridging} & RA-L'22 & RGB+Event & - & - & 23.20 & 60.50 \\
CMX~\cite{zhang2023cmx} & TITS'23 & RGB+Event & 71.88 & 95.64 & 72.42 & 95.07 \\
CMNeXt~\cite{zhang2023delivering} & CVPR'23 & RGB+Event & 72.67 & 95.74 & 72.54 & 95.10 \\
HMNet-B~\cite{hamaguchi2023hierarchical} & CVPR'23 & RGB+Event & - & - & 51.20 & 88.70 \\
HMNet-L~\cite{hamaguchi2023hierarchical} & CVPR'23 & RGB+Event & - & - & 55.00 & 89.80 \\
HALSIE~\cite{das2024halsie} & WACV'24 & RGB+Event & 60.66 & 92.50 & 52.43 & 89.01 \\
EISNet~\cite{xie2024eisnet} & TMM'24 & RGB+Event & 75.03 & 96.04 & 73.07 & 95.12 \\
Hybrid-Segmentation~\cite{li2025efficient} & AAAI'25 & RGB+Event & 67.31 & 95.07 & 66.57 & 94.27 \\
\midrule
Ours & - & RGB+Event & \textbf{77.85} & \textbf{97.10} & \textbf{74.77} & \textbf{95.60} \\
\bottomrule
\end{tabular}
}
\label{tab:ddd17_dsec_results}
\end{table*}

 \begin{tableorg}[tbp]
\centering
\begin{minipage}[t]{0.48\textwidth}
\vspace{0pt}

\centering
\compacttablefont
\caption[Results on MCubeS~\cite{liang2022multimodal}]{Semantic segmentation results on the MCubeS dataset with different modality combinations. ``A'' denotes AoLP, ``D'' denotes DoLP, and ``N'' denotes NIR.}
% \resizebox{\linewidth}{!}{
\fitbox{%
\begin{tabular}{l|c|c|c}
\toprule
\textbf{Method} & \textbf{Venue}  & \textbf{Modal} & \textbf{mIoU (\%)}\\
% \midrule
% Baseline~\cite{cao2024embracing} & ECCV'24 & MiT-B4 & RGB+A+D+N & 43.17 & 60\\
% Our Model & - & MiT-B2 & RGB+A+D+N & 49.70 & 60 \\
% Our Model & - & MiT-B4 & RGB+A+D+N & 49.96 & 60 \\
% Ours & - & MiT-B4 & RGB+A+D+N & \textbf{52.54} & 500 \\
\midrule
MCubeSNet~\cite{liang2022multimodal} & CVPR'22 & RGB+A & 39.10 \\
CMNeXt~\cite{zhang2023delivering} & CVPR'23  & RGB+A & 48.42 \\
% GFBN~\cite{gao2024global} & PR'24 & RGB+A & \textbf{51.56} \\
% ShareCMP~\cite{liu2025sharecmp} & TCSVT'25  & RGB+A & 50.34 \\
Any2Seg~\cite{zheng2024learning} & ECCV'24 & RGB+A & 48.95\\
% MemorySAM~\cite{liao2025memorysammemorizemodalitiessemantics}& Arxiv'25 & RGB+A & 51.2\\
Ours & -  & RGB+A & \textbf{50.09} \\
\midrule
MCubeSNet~\cite{liang2022multimodal} & CVPR'22  & RGB+A+D & 42.00 \\
CMNeXt~\cite{zhang2023delivering} & CVPR'23  & RGB+A+D & 49.48 \\
% GFBN~\cite{gao2024global} & PR'24  & RGB+A+D & \textbf{51.26} \\
ShareCMP~\cite{liu2025sharecmp} & TCSVT'25 & RGB+A+D & 50.99 \\
Any2Seg~\cite{zheng2024learning} & ECCV'24 & RGB+A +D& 50.67\\
Ours & -  & RGB+A+D & \textbf{51.14}  \\
\midrule
FuseNet~\cite{Hazirbas2016FuseNetID} & ACCV’16  & RGB+A+D+N & 40.58  \\
MMTM~\cite{Joze2019MMTMMT} & CVPR'20 & RGB+A+D+N & 39.71  \\
DRConv~\cite{chen2021dynamic} & CVPR'21 & RGB+A+D+N & 34.63 \\
DDF~\cite{zhou2021decoupled} & CVPR'21  & RGB+A+D+N & 36.16  \\
TransFuser~\cite{Prakash2021MultiModalFT} & CVPR'21  & RGB+A+D+N & 37.66 \\
MCubeSNet~\cite{liang2022multimodal} & CVPR'22 & RGB+A+D+N & 42.86  \\
CMNeXt~\cite{zhang2023delivering} & CVPR'23  & RGB+A+D+N & 51.54\\
% GFBN~\cite{gao2024global} & PR'24 & RGB+A+D+N & 52.18 \\
U3M~\cite{li2025u3m} & PR'25  & RGB+A+D+N & 51.69  \\
Any2Seg~\cite{zheng2024learning} & ECCV'24 & RGB+A+D+N & 51.91\\
Ours & -  & RGB+A+D+N & \textbf{52.22} \\
% \midrule
% CMNeXt~\cite{zhang2023delivering} & CVPR'23  & RGB & 48.16 \\
% MCubeSNet~\cite{liang2022multimodal} & CVPR'22  & RGB & 33.70 \\
\bottomrule
\end{tabular}%
}
% }
\label{tab:mcubes_results}

\end{minipage}\hfill
\begin{minipage}[t]{0.48\textwidth}
\vspace{0pt}

\centering
\compacttablefont
\setlength{\tabcolsep}{1pt}
\caption[Results on ZJU~\cite{xiang2021polarization}]{Semantic segmentation results on the ZJU RGB-P dataset. ``P(A+D)'' denotes using both AoLP and DoLP as auxiliary input.}
% \resizebox{\linewidth}{!}{
\fitbox{%
\begin{tabular}{l|c|c|c|c}
\toprule
\textbf{Method} & \textbf{Venue} & \textbf{Backbone} & \textbf{Modal} & \shortstack{\textbf{mIoU}\\\textbf{(\%)}}  \\
\midrule
SwiftNet~\cite{orsic2019defense} & CVPR'19 & - & RGB & 80.3 \\
MultiSenseSeg~\cite{wang2024multisenseseg}  & TGRS'24 & - & RGB & 85.7  \\
% MixImages~\cite{mo2024miximages} & Sensors'24 & - & RGB & 81.9 \\
SegFormer~\cite{xie2021segformer} & NeurIPS'21 & MiT-B2 & RGB & 89.6 \\
% ShareCMP~\cite{liu2025sharecmp} & TCSVT'25 & MiT-B4 & RGB & 92.7 & 500 \\
\midrule
CMX~\cite{zhang2023cmx} & TITS'23 & MiT-B4 & RGB+A & 92.6  \\
% CMX~\cite{zhang2023cmx} & TITS'23 & MiT-B2 & RGB+A & 92.0 & 500 \\
% GFBN~\cite{gao2024global} & PR'24 & MiT-B4 & RGB+A & 92.9  \\
% GFBN~\cite{gao2024global} & PR'24 & MiT-B2 & RGB+A & 92.7 & - \\
Ours & - & MiT-B4 & RGB+A & \textbf{92.8} \\
\midrule
CMX~\cite{zhang2023cmx} & TITS'23 & MiT-B4 & RGB+D & 92.5   \\
% CMX~\cite{zhang2023cmx} & TITS'23 & MiT-B2 & RGB+D & 92.2 & 500 \\
% GFBN~\cite{gao2024global} & PR'24 & MiT-B4 & RGB+D & \textbf{93.1}  \\
% GFBN~\cite{gao2024global} & PR'24 & MiT-B2 & RGB+D & 92.9 & - \\
MultiSenseSeg~\cite{wang2024multisenseseg} & TGRS'24 & - & RGB+D & 86.9  \\
% MixImages~\cite{mo2024miximages} & Sensors'24 & - & RGB+D & 84.8 \\
Ours & - & MiT-B4 & RGB+D & \textbf{93.1} \\
\midrule
RPLENet~\cite{lu2024polarization} & TIV'24 & - & RGB+P(A+D) & 93.1  \\
NLFNet~\cite{yan2021nlfnet} & ROBIO'21 & - & RGB+P(A+D) & 84.4  \\
EAFNet~\cite{xiang2021polarization} & OE'21 & - & RGB+P(A+D)& 85.7 \\
RoadFormer+~\cite{Huang2024RoadFormerDR} & TIV'24 & - & RGB+P(A+D) & 93.0  \\
CAFR~\cite{cao2024embracing} & ECCV'24 & MiT-B4 & RGB+P(A+D)& 83.2 \\
Ours & - & MiT-B4 & RGB+P(A+D)& \textbf{93.8}  \\
\bottomrule
\end{tabular}%
}
% }
\label{tab:zju_rgbp_results}

\end{minipage}
\end{tableorg}

\subsubsection{Results on MCubeS} 
Tab.~\ref{tab:mcubes_results} presents the results on the MCubeS dataset. Using all four modalities (RGB+A+D+N), our method outperforms all compared approaches, including Any2Seg~\cite{zheng2024learning}, U3M~\cite{li2025u3m}, CMNeXt~\cite{zhang2023delivering} and MCubeSNet~\cite{liang2022multimodal}. Experiments with different modality combinations further demonstrate that each additional modality (A, D, and N) contributes to improved segmentation performance. Overall, fusing all four modalities yields the best results. 
% The results for the ZJU dataset are provided in the \textbf{Appendix}.

\subsubsection{Results on ZJU}
On the ZJU dataset, our method achieves 93.8\% mIoU with both A and D inputs, surpassing all competing methods (see Tab.~\ref{tab:zju_rgbp_results}). Under different modality settings, it attains 92.8\% with RGB+A and 93.1\% with RGB+D, outperforming CMX~\cite{zhang2023cmx} in both cases. 
These results highlight the complementary role of polarization cues and demonstrate the effectiveness of our fusion strategy.

\begin{tableorg}[tbp]
\centering
\compacttablefont
\setlength{\tabcolsep}{1pt}
\caption[Results on UrbanLF~\cite{sheng2022urbanlf}]{Semantic segmentation results on the UrbanLF-real and UrbanLF-syn datasets. ``LF80'', ``LF33'', and ``LF8'' denote using 80, 33, and 8 sub-aperture views as auxiliary modalities.}
% \resizebox{\linewidth}{!}{
\fitbox{%
\begin{tabular}{l|c|c|c|c}
\toprule
\multirow{2}{*}{\textbf{Method}} & 
\multirow{2}{*}{\textbf{Venue}} & 
\multirow{2}{*}{\textbf{Modal}} & 
\multicolumn{1}{c|}{\textbf{Real}} & 
\multicolumn{1}{c}{\textbf{Syn}} \\
%\cmidrule(lr){4-5}
\cmidrule(lr){4-5}
 & & & \shortstack{\textbf{mIoU}\\\textbf{(\%)}} & \shortstack{\textbf{mIoU}\\\textbf{(\%)}} \\
 \midrule
PSPNet~\cite{zhao2017pyramid} & CVPR'17  & RGB & 76.34 & 75.78 \\
SegFormer~\cite{xie2021segformer} & NeurIPS'21 & RGB  & 82.20 & 78.53 \\
OCR~\cite{yuan2020object} & TCSVT'22  & RGB & 78.60 & 79.36 \\
\midrule
DAVSS~\cite{zhuang2020video} & TCSVT'20 & Video & 75.91 & 74.27 \\
TMANet~\cite{wang2021temporal} & ICIP' 21  & Video  & 77.14 & 76.41 \\
\midrule
ESANet~\cite{seichter2021efficient} & ICRA'21  & RGB-D  & - & 79.43 \\
SA-Gate~\cite{chen2020bi} & ECCV' 20  & RGB-D & - & 79.53\\
\midrule
PSPNet-LF~\cite{zhao2017pyramid} & CVPR'17 & RGB+LF33  & 78.10 & 77.88 \\
OCR-LF~\cite{sheng2022urbanlf} & TCSVT'22  & RGB+LF33  & 79.32 & 80.43 \\
CMNeXt~\cite{zhang2023delivering} & CVPR'23  & RGB+LF8   & 83.22 & 80.74 \\
CMNeXt~\cite{zhang2023delivering} & CVPR'23  & RGB+LF33  & 82.62 & 80.98 \\
CMNeXt~\cite{zhang2023delivering} & CVPR'23 & RGB+LF80  & 83.11 & 81.02 \\
% \midrule
% LFEI-Net~\cite{jia2025harnessing} & ICASSP'25 & - & LF & - & 83.72 & -\\
% LF-IENet++~\cite{cong2024end} & TCSVT'24 & HRNetV2-W48 & LF & - & - & 80.85 \\
% LF-EANet~\cite{li2024multi} & ICONIP'24 & MiT-B3 & LF & - & - & 80.38 \\
% LFRSNet~\cite{yang2022lfrsnet} & Front. Environ. Sci.'22 & - & LF & - & 79.98 & 80.87 \\
\midrule
% Baseline~\cite{cao2024embracing} & ECCV'24 & MiT-B4 & RGB+LF80 & 60  & 71.25 & 67.05 \\
% Our Model & - & MiT-B4 & RGB+LF80 & 60  & 79.03 & 77.56 \\
Ours & -  & RGB+LF8  & 82.59 & 80.98 \\
Ours & -  & RGB+LF33  & 83.30 & 81.22 \\
Ours & -  & RGB+LF80  & \textbf{83.99} & \textbf{81.91} \\
\bottomrule
\end{tabular}%
}
% }
\label{tab:urbanlf_joint_results}
\end{tableorg}
\subsubsection{Results on UrbanLF}
As shown in Tab.~\ref{tab:urbanlf_joint_results}, our model achieves 83.99\% mIoU on UrbanLF-real and 81.91\% mIoU on UrbanLF-syn, outperforming CMNeXt~\cite{zhang2023delivering} (83.22\%/80.74\%). Even under different sub-aperture configurations, our framework maintains strong performance, achieving 83.30\%/81.22\% with LF33 and 82.59\%/80.98\% with LF8. Compared with RGB-only and video-based methods, our approach delivers substantial improvements, underscoring the advantage of integrating sub-aperture views with RGB guidance.

\Needspace{170pt}
\subsection{Ablation Studies}
% \begin{paracol}{2}
% \switchcolumn
% \end{paracol}

% \begin{table}[t!]
%   \centering
%   \caption{Ablation study of different PLCA branch settings on MFNet. The single branch setting sets the ourput of the opposite branch to zeros, while the duplicated branch setting applies two identical branch. }
%   \label{tab:ablation_plca}
%   \begin{tabular}{lccc}
%     \toprule
%     \textbf{Method} & \textbf{Same-signed} & \textbf{Opposite-signed} & \textbf{mIoU (\%)} \\
%     \midrule
%     Baseline                      & $\times$     & $\times$     & 59.7 \\
%     Same-signed single branch     & $\checkmark$ & $\times$     & 74.5 \\
%     Opposite-signed single branch & $\times$     & $\checkmark$ & 76.5 \\
%     Same-signed duplicated branch & $\checkmark$ & $\times$     & 76.0 \\
%     Full BPLCA                    & $\checkmark$ & $\checkmark$ & \textbf{76.9} \\
%     \bottomrule
%   \end{tabular}
% \end{table}

\Needspace{190pt}
\subsubsection{PLCA Branch Ablation}
\begin{wraptable}{r}{0.48\textwidth}
\centering
\compacttablefont
    \caption{Ablation of PLCA branch settings. `SS` and `OS` denote the number of Same-signed and Opposite-signed branches used. Single: sets opposite to zero; Duplicated: applies two identical branches.}
    \label{tab:ablation_plca}
        \fitbox{%
\begin{tabular}{lccc}
      \toprule
      \textbf{Method} & \textbf{SS} & \textbf{OS} & \textbf{mIoU (\%)} \\
      \midrule
      Baseline        & 0 & 0 & 59.7 \\
      SS Single       & 1 & 0 & 74.5 \\
      OS Single       & 0 & 1 & 76.5 \\
      SS Duplicated   & 2 & 0 & 76.0 \\
      Full BPLCA      & 1 & 1 & \textbf{76.9} \\
      \bottomrule
    \end{tabular}%
}
\end{wraptable}

% To provide quantitative evidence for the PLCA design, we conduct mechanism-level ablations by isolating different branches of BPLCA. To verify that the gain of Full BPLCA stems from complementary opposite-signed interactions rather than increased model capacity, we introduce an SS-Duplicated setting that replaces the opposite-signed branch with a duplicated same-sign branch. 
To provide quantitative evidence for the PLCA design, we conduct mechanism-level ablations by isolating different branches of BPLCA. In addition, We further introduce an SS-Duplicated setting to separate the effect of complementary interactions from increased model capacity. The results on MFNet (RGB+Thermal) are summarized in Tab.~\ref{tab:ablation_plca}. Uni-branch settings lead to a clear performance drop, while combining both branches achieves the best performance, demonstrating that opposite-signed interactions provide complementary information. Qualitative results is provided in Fig.~\ref{fig:pola_vis}.

% \wrapfinish

% \Needspace{210pt}
\subsubsection{Component Ablation}
\begin{inplacetable}
\begin{minipage}[t]{0.48\linewidth}
\centering
\compacttablefont
    \caption[Ablation studies across three datasets]{Ablation study of different architectures on the MFNet and DELIVER (RGB+Depth) datasets.}
    \label{tab:ablation_threecol}
        \fitbox{%
\begin{tabular}{l|c|c}
      \toprule
      \multirow{2}{*}{\textbf{Architecture}} & 
      \textbf{MFNet} & 
      \textbf{DELIVER}\\
      & \textbf{mIoU (\%)} & \textbf{mIoU (\%)} \\
      \midrule
      Baseline 
      & 59.70 & 59.55\\
      + DFCC 
      & 65.40 & 63.54\\
      + BPLCA 
      & 73.66 & 63.70\\
      + BPLCA + DFCC 
      & 76.36 & 64.12\\
      + BPLCA + $\mathcal{L}_{\text{stage}}$ 
      & 76.57 & 63.76\\
      + BPLCA + DFCC & \multirow{2}{*}{\textbf{76.94}} & \multirow{2}{*}{\textbf{64.18}}\\
      \quad \quad\quad\quad\ \ + $\mathcal{L}_\text{stage}$  & \\
      \bottomrule
    \end{tabular}%
}
\end{minipage}
\hfill
\begin{minipage}[t]{0.48\linewidth}
\centering
\compacttablefont
\caption[Ablation studies across three datasets]{Ablation study of different architectures on the DSEC, and DDD17 datasets.}
\label{tab:ablation_2cross}
\fitbox{%
\begin{tabular}{l|c|c}
  \toprule
  \multirow{2}{*}{\textbf{Architecture}} & 
  \textbf{DSEC} & 
  \textbf{DDD17}\\
  & \textbf{mIoU (\%)} & \textbf{mIoU (\%)} \\
  \midrule
  Baseline 
  & 72.42 & 71.88\\
  + DFCC 
  & 72.91 & 74.32 \\
  + BPLCA 
   & 74.04 & 77.57 \\
  + BPLCA + DFCC 
  & 74.29 & 77.73 \\
  + BPLCA + $\mathcal{L}_{\text{stage}}$ 
  & 74.37 & 77.79 \\
  + BPLCA + DFCC &  \multirow{2}{*}{\textbf{74.77}} & 
  \multirow{2}{*}{\textbf{77.85}}\\
  \quad \quad\quad\quad\ \ + $\mathcal{L}_\text{stage}$  &  \\
  \bottomrule
\end{tabular}%
}
\end{minipage}
\end{inplacetable}

% \begin{paracol}{2}
Tab.~\ref{tab:ablation_threecol} summarizes the ablation results on the MFNet and DELIVER (RGB+Depth) datasets, highlighting the contributions of each proposed component in terms of mIoU. % Models are trained for 60 epochs on MFNet and DSEC, and 40 epochs on DDD17. 
Specifically, introducing BPLCA yields substantial gains, demonstrating that BPLCA effectively enhances cross-modal interaction.
Incorporating DFCC further boosts performance by refining feature consistency. The dense stage-wise loss $\mathcal{L}_{stage}$ provides additional supervision and stabilizes optimization, leading to further improvements across all datasets. Finally, combining BPLCA, DFCC, and $\mathcal{L}_{stage}$ achieves the best performance. Overall, these results confirm that the three components are complementary: BPLCA strengthens modality fusion, DFCC enforces feature consistency, and $\mathcal{L}_{\text{stage}}$ supplies multi-level supervision—together delivering significant segmentation improvements. 
% Further results on DDD17 and DSEC datasets are provided in Tab.~\ref{tab:ablation_2cross}. 
 
% \switchcolumn
% \begin{table}[t!]
% \centering
% \caption[Ablation studies across three datasets]{Ablation study of different architectures on the MFNet, DSEC, and DDD17 datasets.}
% \resizebox{0.5\linewidth}{!}{
% \begin{tabular}{l|c|c|c}
% \toprule
% \multirow{2}{*}{\textbf{Architecture}} & 
% \textbf{MFNet} & 
% \textbf{DSEC} & 
% \textbf{DDD17} \\
% %\cmidrule(lr){2-4}
% & \textbf{mIoU (\%)} & \textbf{mIoU (\%)} & \textbf{mIoU (\%)} \\
% \midrule
% % Baseline 
% % & 67.21 & 74.26 & 76.91 \\
% Baseline 
% & 59.7 & 72.42 & 71.88 \\
% +  BPLCA 
% & 73.66 & 74.04 & 77.57 \\
% +  BPLCA + DFCC 
% & 76.36 & 74.29 & 77.73 \\
% + BPLCA + $\mathcal{L}_{\text{stage}}$ 
% & 76.57 & 74.37 & 77.79 \\
% + BPLCA + DFCC + $\mathcal{L}_\text{stage}$  
% & \textbf{76.94} & \textbf{74.77} & \textbf{77.85} \\
% \bottomrule
% \end{tabular}
% }
% \label{tab:ablation_threecol}
% \end{table}
% \end{paracol}

\wrapfinish

We further provide additional ablation results on RGB+Event datasets including DSEC and DDD17 in Tab.~\ref{tab:ablation_2cross}. Certain improvements are observed while combining different components of the proposed method.

\subsection{Efficiency Analysis}\label{sec:efficiency}
 \begin{table*}[hbp]
 \centering
\compacttablefont
  \caption[Computational complexity comparison]{
 GFLOPs and parameter counts of different models using MiT-B2 and MiT-B4 backbones. Lower values indicate better efficiency.
 }
 \renewcommand{\arraystretch}{1.05}
 % Original optional resizebox replaced by the width wrapper below. 
 \setlength{\tabcolsep}{3pt}
 \fitbox{%
\begin{tabular}{l|c|cc|cc}
 \toprule
 \textbf{Model} & \textbf{Input Size} & \multicolumn{2}{c|}{\textbf{MiT-B2}} & \multicolumn{2}{c}{\textbf{MiT-B4}} \\
 \cmidrule(lr){3-6}
 & & \textbf{GFLOPs~$\downarrow$} & \textbf{Params (M)~$\downarrow$} & \textbf{GFLOPs~$\downarrow$} & \textbf{Params (M)~$\downarrow$} \\
 \midrule
 Baseline~\cite{cao2024embracing} & 3$\times$480$\times$640 & 184.8 & \textbf{63.1} & 333.9 & \textbf{136.3} \\
 % CMX~\cite{zhang2023cmx} & 3$\times$480$\times$640 & \textbf{67.6} & 66.6 & \textbf{134.3} & 139.9 \\
 Ours & 3$\times$480$\times$640 & \textbf{162.4} & 66.9 & \textbf{315.1} & 139.8 \\
 \midrule
 Baseline~\cite{cao2024embracing} & 3$\times$512$\times$512 & 152.5 & \textbf{63.1} & 275.6 & \textbf{136.3} \\
 % CMX~\cite{zhang2023cmx} & 3$\times$512$\times$512 & \textbf{65.7} & 66.6 & \textbf{131.1} & 139.9 \\
 Ours & 3$\times$512$\times$512 & \textbf{138.3} & 66.9 & \textbf{264.2} & 139.8 \\
 \bottomrule
 \end{tabular}%
}
 % Original optional resizebox closing delimiter.

 \label{tab:complexity}
 \end{table*}

 \begin{tableorg}[hbp]
 \centering
\compacttablefont
  \caption[Inference speed comparison]{
 Inference speed comparison using MiT-B2 backbone. Higher FPS and lower runtime indicate better efficiency.
 }
 \renewcommand{\arraystretch}{1.05}
 \setlength{\tabcolsep}{3pt}
 \fitbox{ 
 \begin{tabular}{l|c|c|c|c}
 \toprule
 \textbf{Method} & Baseline~\cite{cao2024embracing}  & CMX~\cite{zhang2023cmx} & CMNeXt~\cite{zhang2023delivering}& Ours \\
 \midrule
 FPS~$\uparrow$ & 19.04  & 28.27 & \textbf{30.32} & 16.90\\
 Avg. Time (s)~$\downarrow$ & 0.052 & 0.035 & \textbf{0.033}& 0.059  \\
 \bottomrule
 \end{tabular}
 }

 \label{tab:inference_speed}
 \end{tableorg}

We analyze model efficiency in terms of computational complexity and inference speed. 
As shown in Tab.~\ref{tab:complexity}, our model introduces a modest increase in parameters 
compared with the cross-self-attention~\cite{cao2024embracing} (from 63.1M to 66.9M), mainly due to the 
additional fusion and alignment components. Despite this increase, the proposed design 
reduces overall FLOPs (from 184.8 to 162.4 GFLOPs at $480\times640$), 
indicating that the polarity-aware linear attention and stage-wise fusion avoid the heavy 
quadratic cost typical of standard attention mechanisms.

In terms of runtime efficiency (Tab.~\ref{tab:inference_speed}), our model achieves 16.9 FPS, only slightly lower than 
the cross-self-attention (19.0 FPS), with a small increase in average inference time 
(from 0.052 s to 0.059 s per image). This demonstrates that the added fusion modules incur 
limited practical overhead while enabling consistent performance gains across datasets. 
Overall, the proposed method achieves a favorable trade-off between computational efficiency and 
segmentation accuracy, remaining practical for real-time applications.

\subsection{Qualitative Analysis}
\label{app:qual}

\begin{figure*}[t!]
  \centering
  \includegraphics[width=0.86\textwidth, trim=2.5cm 1.2cm 10cm 0.3cm, clip]{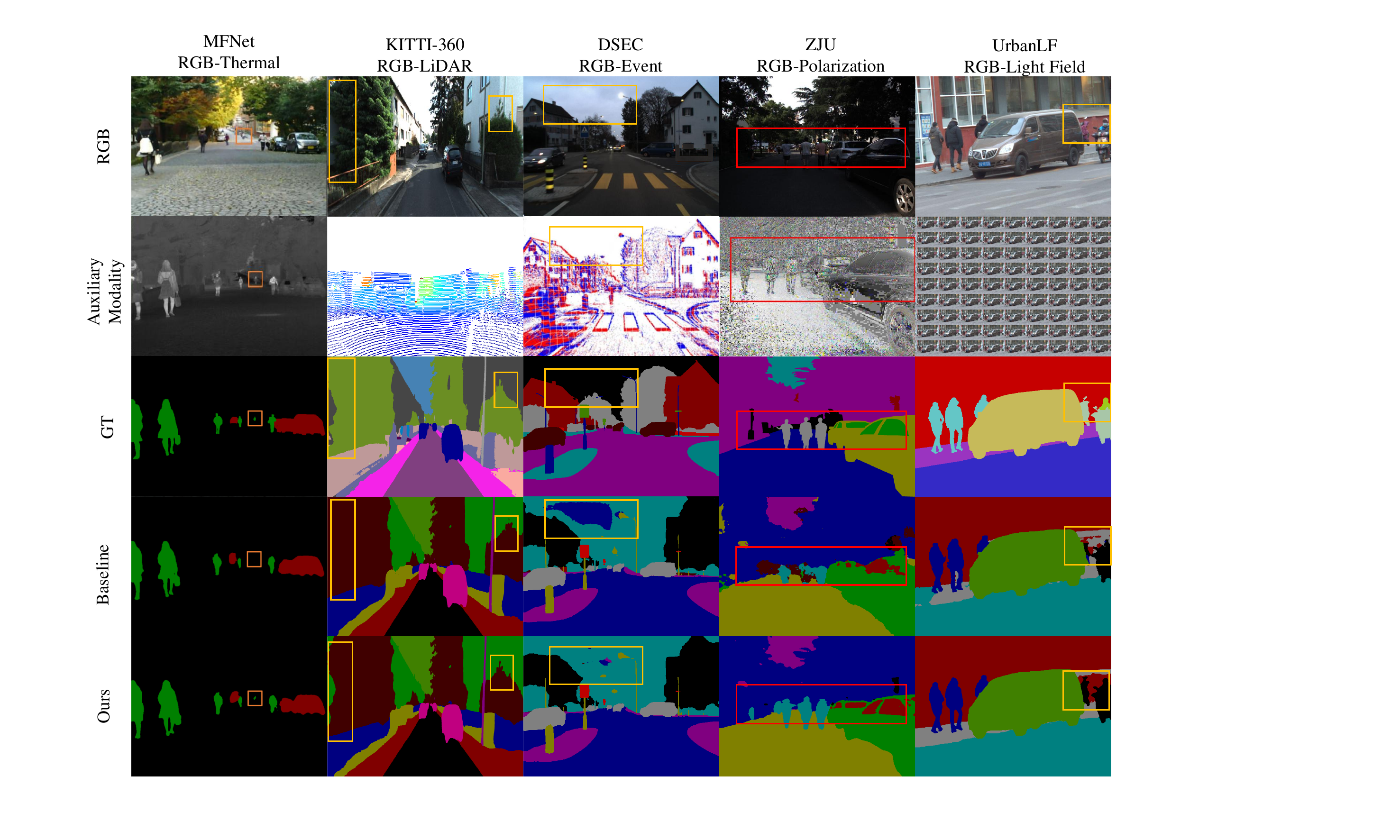}
  \caption{Qualitative results across multiple modalities. Our model consistently produces sharper boundaries and more accurate object recognition under challenging conditions.}
  \label{fig:qualitative}
\end{figure*}

In Fig.~\ref{fig:qualitative}, we present qualitative comparisons on five representative datasets: MFNet~\cite{ha2017mfnet} (RGB-Thermal), KITTI-360~\cite{liao2022kitti} (RGB-LiDAR), DSEC~\cite{gehrig2021dsec} (RGB-Event), ZJU-RGBP~\cite{xiang2021polarization} (RGB-Polarization), and UrbanLF-real~\cite{sheng2022urbanlf} (RGB-LightField). On MFNet, our method more clearly distinguishes pedestrians—particularly small and distant ones—that the baseline often fails to separate from the background. On KITTI-360, it delineates roadside trees with sharper boundaries and smoother transitions between vegetation and surrounding regions. On DSEC, our model suppresses spurious responses in the background sky, yielding more stable predictions under fast-motion conditions. On ZJU-RGBP, improvements are evident around reflective surfaces, where pedestrians and vehicle windows are predicted with higher accuracy and fewer artifacts. On UrbanLF-real, our approach provides more consistent segmentation of riders, pedestrians, and background elements, maintaining coherent parsing even under occlusion. Across all benchmarks, these qualitative visualizations show that our fusion strategy consistently produces more reliable results, underscoring its superiority across diverse sensing scenarios.

\begin{figure*}[t]
  \centering
  \includegraphics[width=0.70\textwidth]{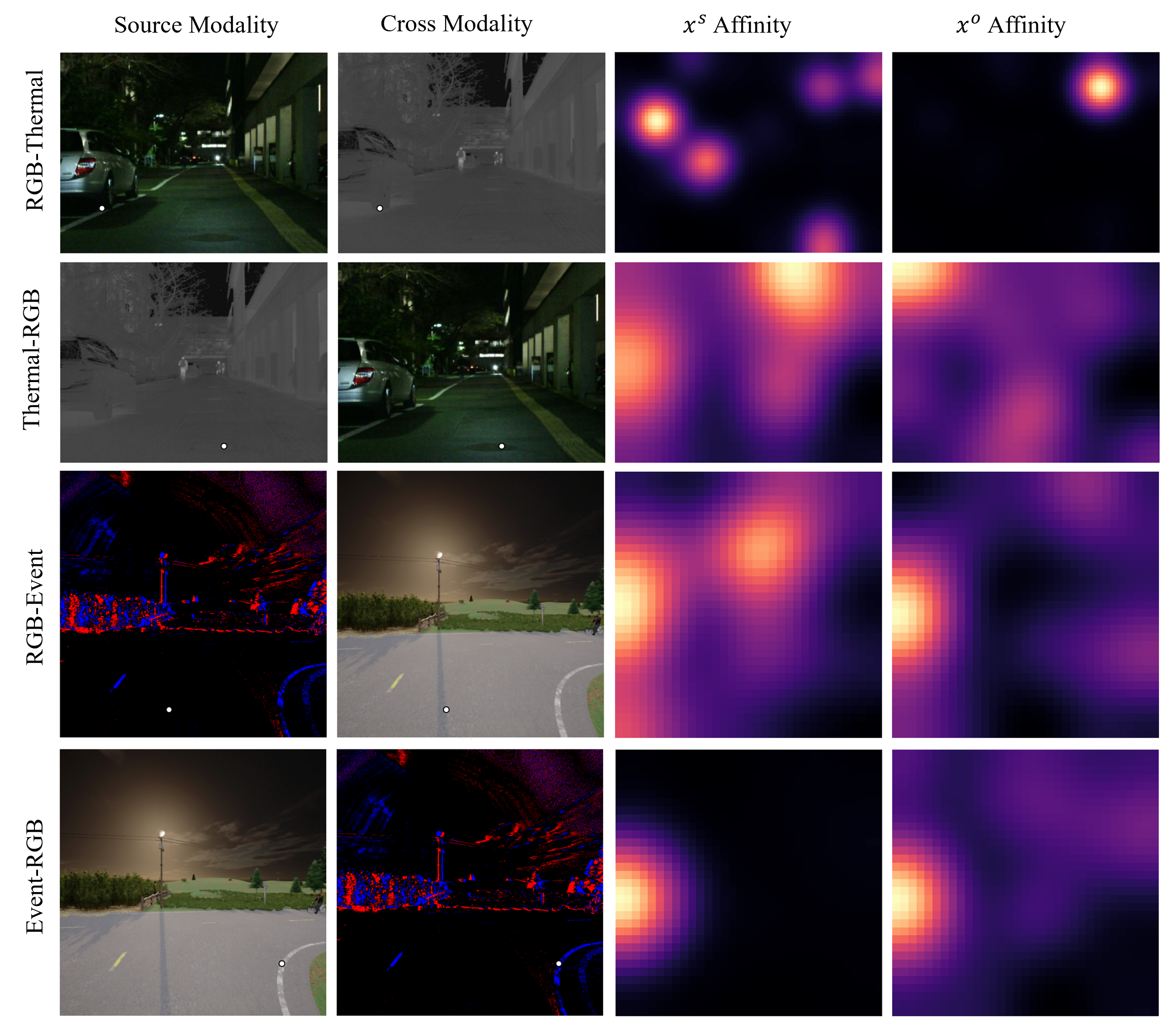}
\caption[BPLCA component visualization]{BPLCA component visualization. RGB-Thermal sample selected from MFNet dataset~\cite{ha2017mfnet}  while RGB-Event sample from DELIVER dataset~\cite{zhang2023delivering}. }
  \label{fig:pola_vis}
\end{figure*}

In Fig.~\ref{fig:pola_vis} we present visualization of the BPLCA component by computing the affinity of the $x^s$ and $x^o$ vector on specific points on the image. We can observe a certain complementary information focus and collaboration between the separated branches.

\subsection{Analysis of ScoreNet Dynamic Modality Selection}
\label{app:score}
% \subsection{Interpretation of ScoreNet.}
% ScoreNet serves as a dynamic auxiliary-modality router in our framework. Instead of treating all auxiliary modalities as equally reliable at every location, it predicts a patch-wise informativeness score for each modality and selects the most discriminative auxiliary embedding at each spatial position. For example, depth cues are usually more informative in structured regions with clear geometric layouts, event features are more sensitive to motion-induced intensity changes and object boundaries, while LiDAR provides sparse but reliable geometric evidence in distant or textureless regions.

% The visualized decision heatmaps validate this behavior. The selected modality patterns vary across both spatial locations and encoder stages. Early-stage maps mainly capture local textures and edges, whereas deeper-stage maps become more coherent and semantically structured. This indicates that ScoreNet performs progressive modality routing along the feature hierarchy. Importantly, the routing maps of different modalities show complementary responses rather than identical activations, confirming that ScoreNet learns adaptive selection instead of static routing. 

\begin{figure*}[t]
  \centering
  \includegraphics[width=0.76\textwidth]{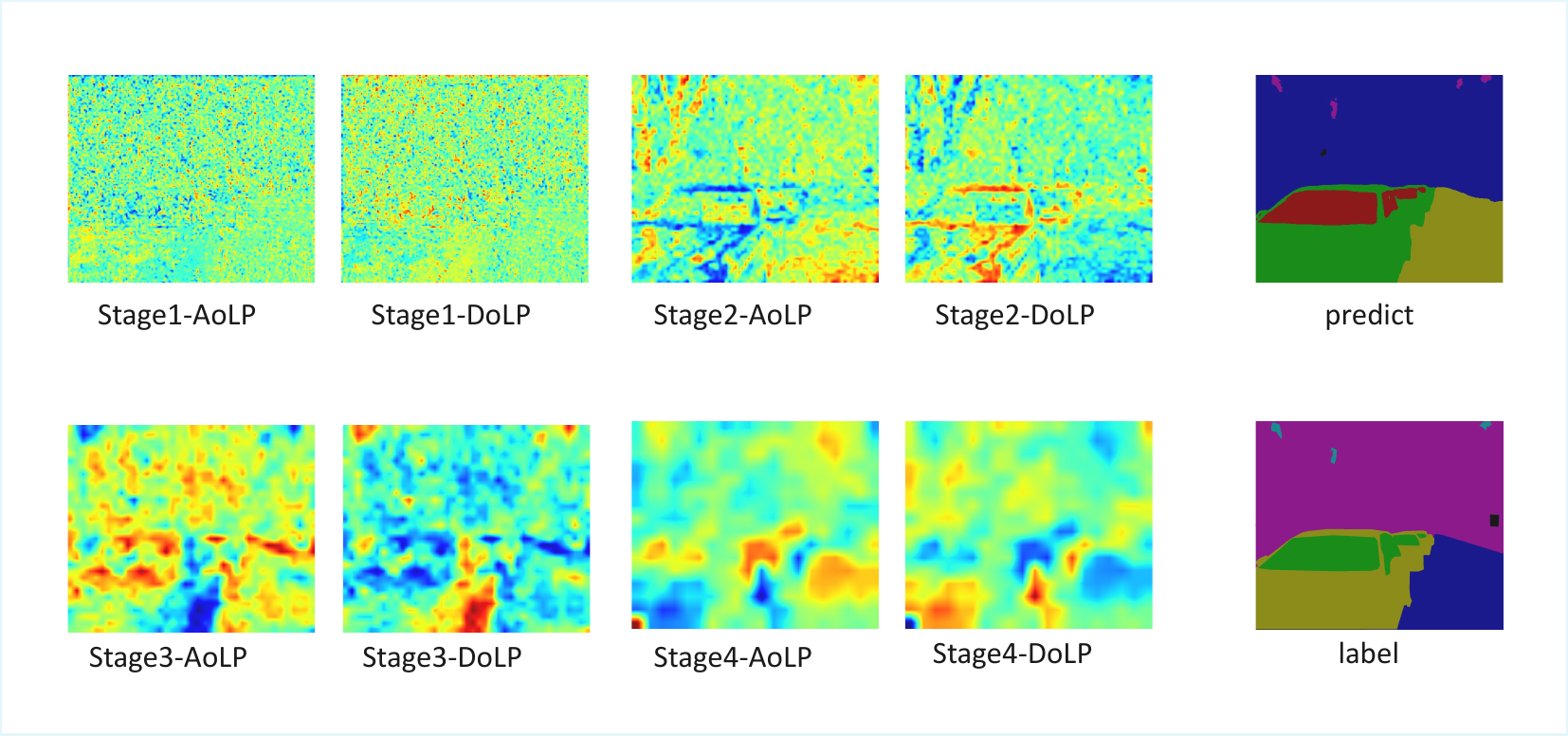}
  \caption[ScoreNet Heatmap on ZJU dataset]{Here is the visualization of ScoreNet on ZJU dataset~\cite{xiang2021polarization}. Red regions indicate higher modality selection scores, while blue regions indicate lower scores.}
  \label{fig:scorenet}
\end{figure*}

To further investigate whether ScoreNet performs meaningful dynamic modality selection, we visualize its decision heatmaps at different encoder stages. ScoreNet assigns an informativeness score to each auxiliary modality at every spatial patch and selects the modality with the highest score to construct a unified auxiliary representation before stage-wise RGB-auxiliary fusion. Therefore, the routing decision is spatially adaptive rather than image-wise or modality-wise fixed.

As shown in Fig.\ref{fig:scorenet}, the ScoreNet responses exhibit clear stage-dependent and modality-dependent patterns. In shallow stages, the score maps contain more fine-grained and high-frequency responses, indicating that early routing mainly focuses on local texture, boundary, and low-level appearance variations. In contrast, the deeper stages show smoother and more semantically organized activation regions, where high-score responses are aligned with object-level structures such as vehicles, road boundaries, and background regions. This suggests that ScoreNet progressively shifts from low-level local selection to high-level semantic modality routing.

Moreover, the decision maps of different auxiliary modalities are not identical. For example, AoLP and DoLP exhibit complementary high-response regions across the scene, indicating that ScoreNet does not collapse to a static preference for a single modality. Instead, it selects different auxiliary cues according to local scene characteristics and feature hierarchy. These observations support our design motivation: ScoreNet first constructs an informative auxiliary feature stream through patch-wise dynamic modality selection, and the subsequent BPCF modules further fuse this selected auxiliary representation with RGB features in a stage-wise manner. The resulting predictions are well aligned with the ground-truth segmentation, demonstrating that the proposed selection mechanism provides meaningful complementary cues for robust semantic segmentation.

%% file: sections/05_conclusion.tex
\section{Conclusion}

In this work, we presented a multi-modal semantic segmentation framework that flexibly integrates RGB with diverse auxiliary modalities, including thermal, LiDAR, depth, event, light field, and polarization data. Our approach features a selective fusion mechanism that dynamically activates the most informative auxiliary modality at each spatial location, a bidirectional polarity-aware linear cross-attention (BPLCA) combined with a dual feature consistency constraint (DFCC) for feature-aligned fusion, and a stage-wise supervision loss that progressively enforces cross-modal consistency. Extensive experiments on eight public benchmarks, including MFNet, KITTI-360, DELIVER, DDD17, DSEC, MCubeS, ZJU, and UrbanLF, demonstrate that our method consistently outperforms SOTA approaches.
% Moreover, qualitative results further validate its effectiveness under challenging conditions such as low light and motion blur.

%% file: main.bbl

\begin{thebibliography}{77}
\ifx \bisbn   \undefined \def \bisbn  #1{ISBN #1}\fi
\ifx \binits  \undefined \def \binits#1{#1}\fi
\ifx \bauthor  \undefined \def \bauthor#1{#1}\fi
\ifx \batitle  \undefined \def \batitle#1{#1}\fi
\ifx \bjtitle  \undefined \def \bjtitle#1{#1}\fi
\ifx \bvolume  \undefined \def \bvolume#1{\textbf{#1}}\fi
\ifx \byear  \undefined \def \byear#1{#1}\fi
\ifx \bissue  \undefined \def \bissue#1{#1}\fi
\ifx \bfpage  \undefined \def \bfpage#1{#1}\fi
\ifx \blpage  \undefined \def \blpage #1{#1}\fi
\ifx \burl  \undefined \def \burl#1{\textsf{#1}}\fi
\ifx \doiurl  \undefined \def \doiurl#1{\url{https://doi.org/#1}}\fi
\ifx \betal  \undefined \def \betal{\textit{et al.}}\fi
\ifx \binstitute  \undefined \def \binstitute#1{#1}\fi
\ifx \binstitutionaled  \undefined \def \binstitutionaled#1{#1}\fi
\ifx \bctitle  \undefined \def \bctitle#1{#1}\fi
\ifx \beditor  \undefined \def \beditor#1{#1}\fi
\ifx \bpublisher  \undefined \def \bpublisher#1{#1}\fi
\ifx \bbtitle  \undefined \def \bbtitle#1{#1}\fi
\ifx \bedition  \undefined \def \bedition#1{#1}\fi
\ifx \bseriesno  \undefined \def \bseriesno#1{#1}\fi
\ifx \blocation  \undefined \def \blocation#1{#1}\fi
\ifx \bsertitle  \undefined \def \bsertitle#1{#1}\fi
\ifx \bsnm \undefined \def \bsnm#1{#1}\fi
\ifx \bsuffix \undefined \def \bsuffix#1{#1}\fi
\ifx \bparticle \undefined \def \bparticle#1{#1}\fi
\ifx \barticle \undefined \def \barticle#1{#1}\fi
\bibcommenthead
\ifx \bconfdate \undefined \def \bconfdate #1{#1}\fi
\ifx \botherref \undefined \def \botherref #1{#1}\fi
\ifx \url \undefined \def \url#1{\textsf{#1}}\fi
\ifx \bchapter \undefined \def \bchapter#1{#1}\fi
\ifx \bbook \undefined \def \bbook#1{#1}\fi
\ifx \bcomment \undefined \def \bcomment#1{#1}\fi
\ifx \oauthor \undefined \def \oauthor#1{#1}\fi
\ifx \citeauthoryear \undefined \def \citeauthoryear#1{#1}\fi
\ifx \endbibitem  \undefined \def \endbibitem {}\fi
\ifx \bconflocation  \undefined \def \bconflocation#1{#1}\fi
\ifx \arxivurl  \undefined \def \arxivurl#1{\textsf{#1}}\fi
\csname PreBibitemsHook\endcsname

%%% 1
\bibitem[\protect\citeauthoryear{Bijelic et~al.}{2020}]{Bijelic2019SeeingTF}
\begin{bchapter}
\bauthor{\bsnm{Bijelic}, \binits{M.}},
\bauthor{\bsnm{Gruber}, \binits{T.}},
\bauthor{\bsnm{Mannan}, \binits{F.}},
\bauthor{\bsnm{Kraus}, \binits{F.}},
\bauthor{\bsnm{Ritter}, \binits{W.}},
\bauthor{\bsnm{Dietmayer}, \binits{K.}},
\bauthor{\bsnm{Heide}, \binits{F.}}:
\bctitle{Seeing through fog without seeing fog: Deep multimodal sensor fusion in unseen adverse weather}.
In: \bbtitle{2020 {IEEE/CVF} Conference on Computer Vision and Pattern Recognition (CVPR)},
pp. \bfpage{11679}--\blpage{11689}
(\byear{2020})
\end{bchapter}
\endbibitem

%%% 2
\bibitem[\protect\citeauthoryear{Zhang et~al.}{2023}]{zhang2023delivering}
\begin{bchapter}
\bauthor{\bsnm{Zhang}, \binits{J.}},
\bauthor{\bsnm{Liu}, \binits{R.}},
\bauthor{\bsnm{Shi}, \binits{H.}},
\bauthor{\bsnm{Yang}, \binits{K.}},
\bauthor{\bsnm{Rei{\ss}}, \binits{S.}},
\bauthor{\bsnm{Peng}, \binits{K.}},
\bauthor{\bsnm{Fu}, \binits{H.}},
\bauthor{\bsnm{Wang}, \binits{K.}},
\bauthor{\bsnm{Stiefelhagen}, \binits{R.}}:
\bctitle{Delivering arbitrary-modal semantic segmentation}.
In: \bbtitle{{IEEE/CVF} Conference on Computer Vision and Pattern Recognition (CVPR)},
pp. \bfpage{1136}--\blpage{1147}
(\byear{2023})
\end{bchapter}
\endbibitem

%%% 3
\bibitem[\protect\citeauthoryear{Chen et~al.}{2020}]{Chen2020EventBasedNV}
\begin{barticle}
\bauthor{\bsnm{Chen}, \binits{G.}},
\bauthor{\bsnm{Cao}, \binits{H.}},
\bauthor{\bsnm{Conradt}, \binits{J.}},
\bauthor{\bsnm{Tang}, \binits{H.}},
\bauthor{\bsnm{R{\"{o}}hrbein}, \binits{F.}},
\bauthor{\bsnm{Knoll}, \binits{A.C.}}:
\batitle{Event-based neuromorphic vision for autonomous driving: {A} paradigm shift for bio-inspired visual sensing and perception}.
\bjtitle{{IEEE} Signal Process. Mag.}
\bvolume{37}(\bissue{4}),
\bfpage{34}--\blpage{49}
(\byear{2020})
\end{barticle}
\endbibitem

%%% 4
\bibitem[\protect\citeauthoryear{Zhang et~al.}{2023}]{zhang2023cmx}
\begin{barticle}
\bauthor{\bsnm{Zhang}, \binits{J.}},
\bauthor{\bsnm{Liu}, \binits{H.}},
\bauthor{\bsnm{Yang}, \binits{K.}},
\bauthor{\bsnm{Hu}, \binits{X.}},
\bauthor{\bsnm{Liu}, \binits{R.}},
\bauthor{\bsnm{Stiefelhagen}, \binits{R.}}:
\batitle{{CMX:} cross-modal fusion for {RGB-X} semantic segmentation with transformers}.
\bjtitle{{IEEE} Trans. Intell. Transp. Syst.}
\bvolume{24}(\bissue{12}),
\bfpage{14679}--\blpage{14694}
(\byear{2023})
\end{barticle}
\endbibitem

%%% 5
\bibitem[\protect\citeauthoryear{Cao et~al.}{2024}]{cao2024embracing}
\begin{bchapter}
\bauthor{\bsnm{Cao}, \binits{H.}},
\bauthor{\bsnm{Zhang}, \binits{Z.}},
\bauthor{\bsnm{Xia}, \binits{Y.}},
\bauthor{\bsnm{Li}, \binits{X.}},
\bauthor{\bsnm{Xia}, \binits{J.}},
\bauthor{\bsnm{Chen}, \binits{G.}},
\bauthor{\bsnm{Knoll}, \binits{A.}}:
\bctitle{Embracing events and frames with hierarchical feature refinement network for object detection}.
In: \bbtitle{European Conference on Computer Vision (ECCV)},
vol. \bseriesno{15142},
pp. \bfpage{161}--\blpage{177}
(\byear{2024})
\end{bchapter}
\endbibitem

%%% 6
\bibitem[\protect\citeauthoryear{Long et~al.}{2015}]{Shelhamer2014FullyCN}
\begin{bchapter}
\bauthor{\bsnm{Long}, \binits{J.}},
\bauthor{\bsnm{Shelhamer}, \binits{E.}},
\bauthor{\bsnm{Darrell}, \binits{T.}}:
\bctitle{Fully convolutional networks for semantic segmentation}.
In: \bbtitle{{IEEE} Conference on Computer Vision and Pattern Recognition (CVPR)},
pp. \bfpage{3431}--\blpage{3440}
(\byear{2015})
\end{bchapter}
\endbibitem

%%% 7
\bibitem[\protect\citeauthoryear{Ronneberger et~al.}{2015}]{Ronneberger2015UNetCN}
\begin{bchapter}
\bauthor{\bsnm{Ronneberger}, \binits{O.}},
\bauthor{\bsnm{Fischer}, \binits{P.}},
\bauthor{\bsnm{Brox}, \binits{T.}}:
\bctitle{U-net: Convolutional networks for biomedical image segmentation}.
In: \beditor{\bsnm{Navab}, \binits{N.}},
\beditor{\bsnm{Hornegger}, \binits{J.}},
\beditor{\bsnm{III}, \binits{W.M.W.}},
\beditor{\bsnm{Frangi}, \binits{A.F.}} (eds.)
\bbtitle{Medical Image Computing and Computer-Assisted Intervention (MICCAI)},
vol. \bseriesno{9351},
pp. \bfpage{234}--\blpage{241}
(\byear{2015})
\end{bchapter}
\endbibitem

%%% 8
\bibitem[\protect\citeauthoryear{Badrinarayanan et~al.}{2017}]{Badrinarayanan2015SegNetAD}
\begin{barticle}
\bauthor{\bsnm{Badrinarayanan}, \binits{V.}},
\bauthor{\bsnm{Kendall}, \binits{A.}},
\bauthor{\bsnm{Cipolla}, \binits{R.}}:
\batitle{Segnet: {A} deep convolutional encoder-decoder architecture for image segmentation}.
\bjtitle{{IEEE} Trans. Pattern Anal. Mach. Intell.}
\bvolume{39}(\bissue{12}),
\bfpage{2481}--\blpage{2495}
(\byear{2017})
\end{barticle}
\endbibitem

%%% 9
\bibitem[\protect\citeauthoryear{Chen et~al.}{2018}]{Chen2018EncoderDecoderWA}
\begin{bchapter}
\bauthor{\bsnm{Chen}, \binits{L.}},
\bauthor{\bsnm{Zhu}, \binits{Y.}},
\bauthor{\bsnm{Papandreou}, \binits{G.}},
\bauthor{\bsnm{Schroff}, \binits{F.}},
\bauthor{\bsnm{Adam}, \binits{H.}}:
\bctitle{Encoder-decoder with atrous separable convolution for semantic image segmentation}.
In: \beditor{\bsnm{Ferrari}, \binits{V.}},
\beditor{\bsnm{Hebert}, \binits{M.}},
\beditor{\bsnm{Sminchisescu}, \binits{C.}},
\beditor{\bsnm{Weiss}, \binits{Y.}} (eds.)
\bbtitle{European Conference on Computer Vision (ECCV)},
vol. \bseriesno{11211},
pp. \bfpage{833}--\blpage{851}
(\byear{2018})
\end{bchapter}
\endbibitem

%%% 10
\bibitem[\protect\citeauthoryear{Dosovitskiy et~al.}{2021}]{Dosovitskiy2020AnII}
\begin{bchapter}
\bauthor{\bsnm{Dosovitskiy}, \binits{A.}},
\bauthor{\bsnm{Beyer}, \binits{L.}},
\bauthor{\bsnm{Kolesnikov}, \binits{A.}},
\bauthor{\bsnm{Weissenborn}, \binits{D.}},
\bauthor{\bsnm{Zhai}, \binits{X.}},
\bauthor{\bsnm{Unterthiner}, \binits{T.}},
\bauthor{\bsnm{Dehghani}, \binits{M.}},
\bauthor{\bsnm{Minderer}, \binits{M.}},
\bauthor{\bsnm{Heigold}, \binits{G.}},
\bauthor{\bsnm{Gelly}, \binits{S.}},
\bauthor{\bsnm{Uszkoreit}, \binits{J.}},
\bauthor{\bsnm{Houlsby}, \binits{N.}}:
\bctitle{An image is worth 16x16 words: Transformers for image recognition at scale}.
In: \bbtitle{9th International Conference on Learning Representations (ICLR)}
(\byear{2021})
\end{bchapter}
\endbibitem

%%% 11
\bibitem[\protect\citeauthoryear{Liu et~al.}{2021}]{Liu2021SwinTH}
\begin{bchapter}
\bauthor{\bsnm{Liu}, \binits{Z.}},
\bauthor{\bsnm{Lin}, \binits{Y.}},
\bauthor{\bsnm{Cao}, \binits{Y.}},
\bauthor{\bsnm{Hu}, \binits{H.}},
\bauthor{\bsnm{Wei}, \binits{Y.}},
\bauthor{\bsnm{Zhang}, \binits{Z.}},
\bauthor{\bsnm{Lin}, \binits{S.}},
\bauthor{\bsnm{Guo}, \binits{B.}}:
\bctitle{Swin transformer: Hierarchical vision transformer using shifted windows}.
In: \bbtitle{2021 {IEEE/CVF} International Conference on Computer Vision (ICCV)},
pp. \bfpage{9992}--\blpage{10002}
(\byear{2021})
\end{bchapter}
\endbibitem

%%% 12
\bibitem[\protect\citeauthoryear{Xie et~al.}{2021}]{xie2021segformer}
\begin{bchapter}
\bauthor{\bsnm{Xie}, \binits{E.}},
\bauthor{\bsnm{Wang}, \binits{W.}},
\bauthor{\bsnm{Yu}, \binits{Z.}},
\bauthor{\bsnm{Anandkumar}, \binits{A.}},
\bauthor{\bsnm{{\'{A}}lvarez}, \binits{J.M.}},
\bauthor{\bsnm{Luo}, \binits{P.}}:
\bctitle{Segformer: Simple and efficient design for semantic segmentation with transformers}.
In: \bbtitle{Annual Conference on Neural Information Processing Systems (NeurIPS)},
pp. \bfpage{12077}--\blpage{12090}
(\byear{2021})
\end{bchapter}
\endbibitem

%%% 13
\bibitem[\protect\citeauthoryear{Cao et~al.}{2024}]{Cao2024SDPTSD}
\begin{barticle}
\bauthor{\bsnm{Cao}, \binits{H.}},
\bauthor{\bsnm{Chen}, \binits{G.}},
\bauthor{\bsnm{Zhao}, \binits{H.}},
\bauthor{\bsnm{Jiang}, \binits{D.}},
\bauthor{\bsnm{Zhang}, \binits{X.}},
\bauthor{\bsnm{Tian}, \binits{Q.}},
\bauthor{\bsnm{Knoll}, \binits{A.}}:
\batitle{{SDPT:} semantic-aware dimension-pooling transformer for image segmentation}.
\bjtitle{{IEEE} Trans. Intell. Transp. Syst.}
\bvolume{25}(\bissue{11}),
\bfpage{15934}--\blpage{15946}
(\byear{2024})
\end{barticle}
\endbibitem

%%% 14
\bibitem[\protect\citeauthoryear{Sun et~al.}{2022}]{Sun2021EventBasedFF}
\begin{bchapter}
\bauthor{\bsnm{Sun}, \binits{L.}},
\bauthor{\bsnm{Sakaridis}, \binits{C.}},
\bauthor{\bsnm{Liang}, \binits{J.}},
\bauthor{\bsnm{Jiang}, \binits{Q.}},
\bauthor{\bsnm{Yang}, \binits{K.}},
\bauthor{\bsnm{Sun}, \binits{P.}},
\bauthor{\bsnm{Ye}, \binits{Y.}},
\bauthor{\bsnm{Wang}, \binits{K.}},
\bauthor{\bsnm{Gool}, \binits{L.V.}}:
\bctitle{Event-based fusion for motion deblurring with cross-modal attention}.
In: \bbtitle{European Conference on Computer Vision (ECCV)},
vol. \bseriesno{13678},
pp. \bfpage{412}--\blpage{428}
(\byear{2022})
\end{bchapter}
\endbibitem

%%% 15
\bibitem[\protect\citeauthoryear{Zhou et~al.}{2023}]{Zhou2022RGBEventFF}
\begin{bchapter}
\bauthor{\bsnm{Zhou}, \binits{Z.}},
\bauthor{\bsnm{Wu}, \binits{Z.}},
\bauthor{\bsnm{Boutteau}, \binits{R.}},
\bauthor{\bsnm{Yang}, \binits{F.}},
\bauthor{\bsnm{Demonceaux}, \binits{C.}},
\bauthor{\bsnm{Ginhac}, \binits{D.}}:
\bctitle{Rgb-event fusion for moving object detection in autonomous driving}.
In: \bbtitle{{IEEE} International Conference on Robotics and Automation (ICRA)},
pp. \bfpage{7808}--\blpage{7815}
(\byear{2023})
\end{bchapter}
\endbibitem

%%% 16
\bibitem[\protect\citeauthoryear{Xie et~al.}{2024}]{xie2024eisnet}
\begin{barticle}
\bauthor{\bsnm{Xie}, \binits{B.}},
\bauthor{\bsnm{Deng}, \binits{Y.}},
\bauthor{\bsnm{Shao}, \binits{Z.}},
\bauthor{\bsnm{Li}, \binits{Y.}}:
\batitle{Eisnet: {A} multi-modal fusion network for semantic segmentation with events and images}.
\bjtitle{{IEEE} Trans. Multim.}
\bvolume{26},
\bfpage{8639}--\blpage{8650}
(\byear{2024})
\end{barticle}
\endbibitem

%%% 17
\bibitem[\protect\citeauthoryear{Cao et~al.}{2022}]{9787342}
\begin{barticle}
\bauthor{\bsnm{Cao}, \binits{H.}},
\bauthor{\bsnm{Chen}, \binits{G.}},
\bauthor{\bsnm{Li}, \binits{Z.}},
\bauthor{\bsnm{Hu}, \binits{Y.}},
\bauthor{\bsnm{Knoll}, \binits{A.}}:
\batitle{Neurograsp: Multimodal neural network with euler region regression for neuromorphic vision-based grasp pose estimation}.
\bjtitle{IEEE Transactions on Instrumentation and Measurement}
\bvolume{71},
\bfpage{1}--\blpage{11}
(\byear{2022})
\end{barticle}
\endbibitem

%%% 18
\bibitem[\protect\citeauthoryear{Zheng et~al.}{2024}]{zheng2024learning}
\begin{bchapter}
\bauthor{\bsnm{Zheng}, \binits{X.}},
\bauthor{\bsnm{Lyu}, \binits{Y.}},
\bauthor{\bsnm{Wang}, \binits{L.}}:
\bctitle{Learning modality-agnostic representation for semantic segmentation from any modalities}.
In: \beditor{\bsnm{Leonardis}, \binits{A.}},
\beditor{\bsnm{Ricci}, \binits{E.}},
\beditor{\bsnm{Roth}, \binits{S.}},
\beditor{\bsnm{Russakovsky}, \binits{O.}},
\beditor{\bsnm{Sattler}, \binits{T.}},
\beditor{\bsnm{Varol}, \binits{G.}} (eds.)
\bbtitle{European Conference on Computer Vision (ECCV)},
vol. \bseriesno{15092},
pp. \bfpage{146}--\blpage{165}
(\byear{2024})
\end{bchapter}
\endbibitem

%%% 19
\bibitem[\protect\citeauthoryear{Li et~al.}{2025}]{li2025efficient}
\begin{bchapter}
\bauthor{\bsnm{Li}, \binits{H.}},
\bauthor{\bsnm{Peng}, \binits{Y.}},
\bauthor{\bsnm{Yuan}, \binits{J.}},
\bauthor{\bsnm{Wu}, \binits{P.}},
\bauthor{\bsnm{Wang}, \binits{J.}},
\bauthor{\bsnm{Zhang}, \binits{Y.}},
\bauthor{\bsnm{Sun}, \binits{X.}}:
\bctitle{Efficient event-based semantic segmentation via exploiting frame-event fusion: {A} hybrid neural network approach}.
In: \beditor{\bsnm{Walsh}, \binits{T.}},
\beditor{\bsnm{Shah}, \binits{J.}},
\beditor{\bsnm{Kolter}, \binits{Z.}} (eds.)
\bbtitle{Association for the Advancement of Artificial Intelligence (AAAI)},
pp. \bfpage{18296}--\blpage{18304}
(\byear{2025})
\end{bchapter}
\endbibitem

%%% 20
\bibitem[\protect\citeauthoryear{Prakash et~al.}{2021}]{Prakash2021MultiModalFT}
\begin{bchapter}
\bauthor{\bsnm{Prakash}, \binits{A.}},
\bauthor{\bsnm{Chitta}, \binits{K.}},
\bauthor{\bsnm{Geiger}, \binits{A.}}:
\bctitle{Multi-modal fusion transformer for end-to-end autonomous driving}.
In: \bbtitle{{IEEE} Conference on Computer Vision and Pattern Recognition (CVPR)},
pp. \bfpage{7077}--\blpage{7087}
(\byear{2021})
\end{bchapter}
\endbibitem

%%% 21
\bibitem[\protect\citeauthoryear{Joze et~al.}{2020}]{Joze2019MMTMMT}
\begin{bchapter}
\bauthor{\bsnm{Joze}, \binits{H.R.V.}},
\bauthor{\bsnm{Shaban}, \binits{A.}},
\bauthor{\bsnm{Iuzzolino}, \binits{M.L.}},
\bauthor{\bsnm{Koishida}, \binits{K.}}:
\bctitle{{MMTM:} multimodal transfer module for {CNN} fusion}.
In: \bbtitle{2020 {IEEE/CVF} Conference on Computer Vision and Pattern Recognition (CVPR)},
pp. \bfpage{13286}--\blpage{13296}
(\byear{2020})
\end{bchapter}
\endbibitem

%%% 22
\bibitem[\protect\citeauthoryear{Hazirbas et~al.}{2016}]{Hazirbas2016FuseNetID}
\begin{bchapter}
\bauthor{\bsnm{Hazirbas}, \binits{C.}},
\bauthor{\bsnm{Ma}, \binits{L.}},
\bauthor{\bsnm{Domokos}, \binits{C.}},
\bauthor{\bsnm{Cremers}, \binits{D.}}:
\bctitle{Fusenet: Incorporating depth into semantic segmentation via fusion-based {CNN} architecture}.
In: \beditor{\bsnm{Lai}, \binits{S.}},
\beditor{\bsnm{Lepetit}, \binits{V.}},
\beditor{\bsnm{Nishino}, \binits{K.}},
\beditor{\bsnm{Sato}, \binits{Y.}} (eds.)
\bbtitle{13th Asian Conference on Computer Vision (ACCV)},
vol. \bseriesno{10111},
pp. \bfpage{213}--\blpage{228}
(\byear{2016})
\end{bchapter}
\endbibitem

%%% 23
\bibitem[\protect\citeauthoryear{Cai et~al.}{2025}]{cai2025keep}
\begin{bchapter}
\bauthor{\bsnm{Cai}, \binits{J.}},
\bauthor{\bsnm{Su}, \binits{J.}},
\bauthor{\bsnm{Li}, \binits{Q.}},
\bauthor{\bsnm{Yang}, \binits{W.}},
\bauthor{\bsnm{Wang}, \binits{S.}},
\bauthor{\bsnm{Zhao}, \binits{T.}},
\bauthor{\bsnm{He}, \binits{S.}},
\bauthor{\bsnm{Liu}, \binits{W.}}:
\bctitle{Keep the balance: {A} parameter-efficient symmetrical framework for {RGB+X} semantic segmentation}.
In: \bbtitle{{IEEE/CVF} Conference on Computer Vision and Pattern Recognition (CVPR)},
pp. \bfpage{10587}--\blpage{10598}
(\byear{2025})
\end{bchapter}
\endbibitem

%%% 24
\bibitem[\protect\citeauthoryear{Liang et~al.}{2022}]{liang2022multimodal}
\begin{bchapter}
\bauthor{\bsnm{Liang}, \binits{Y.}},
\bauthor{\bsnm{Wakaki}, \binits{R.}},
\bauthor{\bsnm{Nobuhara}, \binits{S.}},
\bauthor{\bsnm{Nishino}, \binits{K.}}:
\bctitle{Multimodal material segmentation}.
In: \bbtitle{Proceedings of the IEEE/CVF Conference on Computer Vision and Pattern Recognition (CVPR)},
pp. \bfpage{19800}--\blpage{19808}
(\byear{2022})
\end{bchapter}
\endbibitem

%%% 25
\bibitem[\protect\citeauthoryear{Meng et~al.}{2025}]{Meng2025PolaFormerPL}
\begin{bchapter}
\bauthor{\bsnm{Meng}, \binits{W.}},
\bauthor{\bsnm{Luo}, \binits{Y.}},
\bauthor{\bsnm{Li}, \binits{X.}},
\bauthor{\bsnm{Jiang}, \binits{D.}},
\bauthor{\bsnm{Zhang}, \binits{Z.}}:
\bctitle{Polaformer: Polarity-aware linear attention for vision transformers}.
In: \bbtitle{The Thirteenth International Conference on Learning Representations (ICLR)}
(\byear{2025})
\end{bchapter}
\endbibitem

%%% 26
\bibitem[\protect\citeauthoryear{Zheng et~al.}{2024}]{Zheng2024CenteringTV}
\begin{bchapter}
\bauthor{\bsnm{Zheng}, \binits{X.}},
\bauthor{\bsnm{Lyu}, \binits{Y.}},
\bauthor{\bsnm{Zhou}, \binits{J.}},
\bauthor{\bsnm{Wang}, \binits{L.}}:
\bctitle{Centering the value of every modality: Towards efficient and resilient modality-agnostic semantic segmentation}.
In: \bbtitle{European Conference on Computer Vision (ECCV)},
vol. \bseriesno{15127},
pp. \bfpage{192}--\blpage{212}
(\byear{2024})
\end{bchapter}
\endbibitem

%%% 27
\bibitem[\protect\citeauthoryear{Jia et~al.}{2024}]{Jia2024GeminiFusionEP}
\begin{bchapter}
\bauthor{\bsnm{Jia}, \binits{D.}},
\bauthor{\bsnm{Guo}, \binits{J.}},
\bauthor{\bsnm{Han}, \binits{K.}},
\bauthor{\bsnm{Wu}, \binits{H.}},
\bauthor{\bsnm{Zhang}, \binits{C.}},
\bauthor{\bsnm{Xu}, \binits{C.}},
\bauthor{\bsnm{Chen}, \binits{X.}}:
\bctitle{Geminifusion: Efficient pixel-wise multimodal fusion for vision transformer}.
In: \bbtitle{Forty-first International Conference on Machine Learning (ICML)}
(\byear{2024})
\end{bchapter}
\endbibitem

%%% 28
\bibitem[\protect\citeauthoryear{Srivastava and Sharma}{2024}]{Srivastava2024OmniVec2A}
\begin{bchapter}
\bauthor{\bsnm{Srivastava}, \binits{S.}},
\bauthor{\bsnm{Sharma}, \binits{G.}}:
\bctitle{Omnivec2-a novel transformer based network for large scale multimodal and multitask learning}.
In: \bbtitle{Proceedings of the IEEE/CVF Conference on Computer Vision and Pattern Recognition (CVPR)},
pp. \bfpage{27412}--\blpage{27424}
(\byear{2024})
\end{bchapter}
\endbibitem

%%% 29
\bibitem[\protect\citeauthoryear{Katharopoulos et~al.}{2020}]{kat2020linearatt}
\begin{bchapter}
\bauthor{\bsnm{Katharopoulos}, \binits{A.}},
\bauthor{\bsnm{Vyas}, \binits{A.}},
\bauthor{\bsnm{Pappas}, \binits{N.}},
\bauthor{\bsnm{Fleuret}, \binits{F.}}:
\bctitle{Transformers are rnns: Fast autoregressive transformers with linear attention}.
In: \bbtitle{Proceedings of the 37th International Conference on Machine Learning (ICML)},
vol. \bseriesno{119},
pp. \bfpage{5156}--\blpage{5165}
(\byear{2020})
\end{bchapter}
\endbibitem

%%% 30
\bibitem[\protect\citeauthoryear{Ha et~al.}{2017}]{ha2017mfnet}
\begin{bchapter}
\bauthor{\bsnm{Ha}, \binits{Q.}},
\bauthor{\bsnm{Watanabe}, \binits{K.}},
\bauthor{\bsnm{Karasawa}, \binits{T.}},
\bauthor{\bsnm{Ushiku}, \binits{Y.}},
\bauthor{\bsnm{Harada}, \binits{T.}}:
\bctitle{Mfnet: Towards real-time semantic segmentation for autonomous vehicles with multi-spectral scenes}.
In: \bbtitle{2017 {IEEE/RSJ} International Conference on Intelligent Robots and Systems (IROS)},
pp. \bfpage{5108}--\blpage{5115}
(\byear{2017})
\end{bchapter}
\endbibitem

%%% 31
\bibitem[\protect\citeauthoryear{Liao et~al.}{2023}]{liao2022kitti}
\begin{barticle}
\bauthor{\bsnm{Liao}, \binits{Y.}},
\bauthor{\bsnm{Xie}, \binits{J.}},
\bauthor{\bsnm{Geiger}, \binits{A.}}:
\batitle{{KITTI-360:} {A} novel dataset and benchmarks for urban scene understanding in 2d and 3d}.
\bjtitle{{IEEE} Trans. Pattern Anal. Mach. Intell.}
\bvolume{45}(\bissue{3}),
\bfpage{3292}--\blpage{3310}
(\byear{2023})
\end{barticle}
\endbibitem

%%% 32
\bibitem[\protect\citeauthoryear{Binas et~al.}{2017}]{Binas2017DDD17ED}
\begin{botherref}
\oauthor{\bsnm{Binas}, \binits{J.}},
\oauthor{\bsnm{Neil}, \binits{D.}},
\oauthor{\bsnm{Liu}, \binits{S.-C.}},
\oauthor{\bsnm{Delbruck}, \binits{T.}}:
Ddd17: End-to-end davis driving dataset.
arXiv preprint arXiv:1711.01458
(2017)
\end{botherref}
\endbibitem

%%% 33
\bibitem[\protect\citeauthoryear{Gehrig et~al.}{2021}]{gehrig2021dsec}
\begin{barticle}
\bauthor{\bsnm{Gehrig}, \binits{M.}},
\bauthor{\bsnm{Aarents}, \binits{W.}},
\bauthor{\bsnm{Gehrig}, \binits{D.}},
\bauthor{\bsnm{Scaramuzza}, \binits{D.}}:
\batitle{{DSEC:} {A} stereo event camera dataset for driving scenarios}.
\bjtitle{{IEEE} Robotics Autom. Lett.}
\bvolume{6}(\bissue{3}),
\bfpage{4947}--\blpage{4954}
(\byear{2021})
\end{barticle}
\endbibitem

%%% 34
\bibitem[\protect\citeauthoryear{Sheng et~al.}{2022}]{sheng2022urbanlf}
\begin{barticle}
\bauthor{\bsnm{Sheng}, \binits{H.}},
\bauthor{\bsnm{Cong}, \binits{R.}},
\bauthor{\bsnm{Yang}, \binits{D.}},
\bauthor{\bsnm{Chen}, \binits{R.}},
\bauthor{\bsnm{Wang}, \binits{S.}},
\bauthor{\bsnm{Cui}, \binits{Z.}}:
\batitle{Urbanlf: {A} comprehensive light field dataset for semantic segmentation of urban scenes}.
\bjtitle{{IEEE} Trans. Circuits Syst. Video Technol.}
\bvolume{32}(\bissue{11}),
\bfpage{7880}--\blpage{7893}
(\byear{2022})
\end{barticle}
\endbibitem

%%% 35
\bibitem[\protect\citeauthoryear{Xiang et~al.}{2021}]{xiang2021polarization}
\begin{barticle}
\bauthor{\bsnm{Xiang}, \binits{K.}},
\bauthor{\bsnm{Yang}, \binits{K.}},
\bauthor{\bsnm{Wang}, \binits{K.}}:
\batitle{Polarization-driven semantic segmentation via efficient attention-bridged fusion}.
\bjtitle{Optics Express}
\bvolume{29}(\bissue{4}),
\bfpage{4802}--\blpage{4820}
(\byear{2021})
\end{barticle}
\endbibitem

%%% 36
\bibitem[\protect\citeauthoryear{Cordts et~al.}{2016}]{cordts2016cityscapes}
\begin{bchapter}
\bauthor{\bsnm{Cordts}, \binits{M.}},
\bauthor{\bsnm{Omran}, \binits{M.}},
\bauthor{\bsnm{Ramos}, \binits{S.}},
\bauthor{\bsnm{Rehfeld}, \binits{T.}},
\bauthor{\bsnm{Enzweiler}, \binits{M.}},
\bauthor{\bsnm{Benenson}, \binits{R.}},
\bauthor{\bsnm{Franke}, \binits{U.}},
\bauthor{\bsnm{Roth}, \binits{S.}},
\bauthor{\bsnm{Schiele}, \binits{B.}}:
\bctitle{The cityscapes dataset for semantic urban scene understanding}.
In: \bbtitle{2016 {IEEE} Conference on Computer Vision and Pattern Recognition (CVPR)},
pp. \bfpage{3213}--\blpage{3223}
(\byear{2016})
\end{bchapter}
\endbibitem

%%% 37
\bibitem[\protect\citeauthoryear{Alonso and Murillo}{2019}]{alonso2019ev}
\begin{bchapter}
\bauthor{\bsnm{Alonso}, \binits{I.}},
\bauthor{\bsnm{Murillo}, \binits{A.C.}}:
\bctitle{Ev-segnet: Semantic segmentation for event-based cameras}.
In: \bbtitle{{IEEE} Conference on Computer Vision and Pattern Recognition Workshops (CVPRW)},
pp. \bfpage{1624}--\blpage{1633}
(\byear{2019})
\end{bchapter}
\endbibitem

%%% 38
\bibitem[\protect\citeauthoryear{Loshchilov and Hutter}{2019}]{loshchilov2019decoupled}
\begin{bchapter}
\bauthor{\bsnm{Loshchilov}, \binits{I.}},
\bauthor{\bsnm{Hutter}, \binits{F.}}:
\bctitle{Decoupled weight decay regularization}.
In: \bbtitle{7th International Conference on Learning Representations (ICLR)}
(\byear{2019})
\end{bchapter}
\endbibitem

%%% 39
\bibitem[\protect\citeauthoryear{Zhang et~al.}{2019}]{zhang2019pattern}
\begin{bchapter}
\bauthor{\bsnm{Zhang}, \binits{Z.}},
\bauthor{\bsnm{Cui}, \binits{Z.}},
\bauthor{\bsnm{Xu}, \binits{C.}},
\bauthor{\bsnm{Yan}, \binits{Y.}},
\bauthor{\bsnm{Sebe}, \binits{N.}},
\bauthor{\bsnm{Yang}, \binits{J.}}:
\bctitle{Pattern-affinitive propagation across depth, surface normal and semantic segmentation}.
In: \bbtitle{{IEEE} Conference on Computer Vision and Pattern Recognition (CVPR)},
pp. \bfpage{4106}--\blpage{4115}
(\byear{2019})
\end{bchapter}
\endbibitem

%%% 40
\bibitem[\protect\citeauthoryear{Zhang et~al.}{2021}]{zhang2021abmdrnet}
\begin{bchapter}
\bauthor{\bsnm{Zhang}, \binits{Q.}},
\bauthor{\bsnm{Zhao}, \binits{S.}},
\bauthor{\bsnm{Luo}, \binits{Y.}},
\bauthor{\bsnm{Zhang}, \binits{D.}},
\bauthor{\bsnm{Huang}, \binits{N.}},
\bauthor{\bsnm{Han}, \binits{J.}}:
\bctitle{Abmdrnet: Adaptive-weighted bi-directional modality difference reduction network for {RGB-T} semantic segmentation}.
In: \bbtitle{{IEEE} Conference on Computer Vision and Pattern Recognition (CVPR)},
pp. \bfpage{2633}--\blpage{2642}
(\byear{2021})
\end{bchapter}
\endbibitem

%%% 41
\bibitem[\protect\citeauthoryear{Zhou et~al.}{2021}]{zhou2021gmnet}
\begin{barticle}
\bauthor{\bsnm{Zhou}, \binits{W.}},
\bauthor{\bsnm{Liu}, \binits{J.}},
\bauthor{\bsnm{Lei}, \binits{J.}},
\bauthor{\bsnm{Yu}, \binits{L.}},
\bauthor{\bsnm{Hwang}, \binits{J.}}:
\batitle{Gmnet: Graded-feature multilabel-learning network for rgb-thermal urban scene semantic segmentation}.
\bjtitle{{IEEE} Trans. Image Process.}
\bvolume{30},
\bfpage{7790}--\blpage{7802}
(\byear{2021})
\end{barticle}
\endbibitem

%%% 42
\bibitem[\protect\citeauthoryear{Zhou et~al.}{2022}]{zhou2022edge}
\begin{bchapter}
\bauthor{\bsnm{Zhou}, \binits{W.}},
\bauthor{\bsnm{Dong}, \binits{S.}},
\bauthor{\bsnm{Xu}, \binits{C.}},
\bauthor{\bsnm{Qian}, \binits{Y.}}:
\bctitle{Edge-aware guidance fusion network for rgb-thermal scene parsing}.
In: \bbtitle{Thirty-Sixth {AAAI} Conference on Artificial Intelligence (AAAI)},
pp. \bfpage{3571}--\blpage{3579}
(\byear{2022})
\end{bchapter}
\endbibitem

%%% 43
\bibitem[\protect\citeauthoryear{Frigo et~al.}{2022}]{frigo2022doodlenet}
\begin{bchapter}
\bauthor{\bsnm{Frigo}, \binits{O.}},
\bauthor{\bsnm{Martin{-}Gaff{\'{e}}}, \binits{L.}},
\bauthor{\bsnm{Wacongne}, \binits{C.}}:
\bctitle{Doodlenet: Double deeplab enhanced feature fusion for thermal-color semantic segmentation}.
In: \bbtitle{{IEEE/CVF} Conference on Computer Vision and Pattern Recognition Workshops (CVPRW)},
pp. \bfpage{3020}--\blpage{3028}
(\byear{2022})
\end{bchapter}
\endbibitem

%%% 44
\bibitem[\protect\citeauthoryear{Yin et~al.}{2025}]{yin2025omnisegmentor}
\begin{botherref}
\oauthor{\bsnm{Yin}, \binits{B.-W.}},
\oauthor{\bsnm{Cao}, \binits{J.-L.}},
\oauthor{\bsnm{Zhang}, \binits{X.}},
\oauthor{\bsnm{Chen}, \binits{Y.}},
\oauthor{\bsnm{Cheng}, \binits{M.-M.}},
\oauthor{\bsnm{Hou}, \binits{Q.}}:
Omnisegmentor: A flexible multi-modal learning framework for semantic segmentation.
The Thirty-ninth Annual Conference on Neural Information Processing Systems (NeurIPS)
(2025)
\end{botherref}
\endbibitem

%%% 45
\bibitem[\protect\citeauthoryear{Zhuang et~al.}{2021}]{zhuang2021perception}
\begin{bchapter}
\bauthor{\bsnm{Zhuang}, \binits{Z.}},
\bauthor{\bsnm{Li}, \binits{R.}},
\bauthor{\bsnm{Jia}, \binits{K.}},
\bauthor{\bsnm{Wang}, \binits{Q.}},
\bauthor{\bsnm{Li}, \binits{Y.}},
\bauthor{\bsnm{Tan}, \binits{M.}}:
\bctitle{Perception-aware multi-sensor fusion for 3d lidar semantic segmentation}.
In: \bbtitle{2021 {IEEE/CVF} International Conference on Computer Vision (ICCV)},
pp. \bfpage{16260}--\blpage{16270}
(\byear{2021})
\end{bchapter}
\endbibitem

%%% 46
\bibitem[\protect\citeauthoryear{Wang et~al.}{2022}]{wang2022multimodal}
\begin{bchapter}
\bauthor{\bsnm{Wang}, \binits{Y.}},
\bauthor{\bsnm{Chen}, \binits{X.}},
\bauthor{\bsnm{Cao}, \binits{L.}},
\bauthor{\bsnm{Huang}, \binits{W.}},
\bauthor{\bsnm{Sun}, \binits{F.}},
\bauthor{\bsnm{Wang}, \binits{Y.}}:
\bctitle{Multimodal token fusion for vision transformers}.
In: \bbtitle{{IEEE/CVF} Conference on Computer Vision and Pattern Recognition (CVPR)},
pp. \bfpage{12176}--\blpage{12185}
(\byear{2022})
\end{bchapter}
\endbibitem

%%% 47
\bibitem[\protect\citeauthoryear{Br{\"{o}}dermann et~al.}{2023}]{broedermann2023hrfuser}
\begin{bchapter}
\bauthor{\bsnm{Br{\"{o}}dermann}, \binits{T.}},
\bauthor{\bsnm{Sakaridis}, \binits{C.}},
\bauthor{\bsnm{Dai}, \binits{D.}},
\bauthor{\bsnm{Gool}, \binits{L.V.}}:
\bctitle{Hrfuser: {A} multi-resolution sensor fusion architecture for 2d object detection}.
In: \bbtitle{26th {IEEE} International Conference on Intelligent Transportation Systems (ITSC)},
pp. \bfpage{4159}--\blpage{4166}
(\byear{2023})
\end{bchapter}
\endbibitem

%%% 48
\bibitem[\protect\citeauthoryear{Yin et~al.}{2024}]{yin2024dformer}
\begin{bchapter}
\bauthor{\bsnm{Yin}, \binits{B.}},
\bauthor{\bsnm{Zhang}, \binits{X.}},
\bauthor{\bsnm{Li}, \binits{Z.}},
\bauthor{\bsnm{Liu}, \binits{L.}},
\bauthor{\bsnm{Cheng}, \binits{M.}},
\bauthor{\bsnm{Hou}, \binits{Q.}}:
\bctitle{Dformer: Rethinking {RGBD} representation learning for semantic segmentation}.
In: \bbtitle{The Twelfth International Conference on Learning Representations (ICLR)}
(\byear{2024})
\end{bchapter}
\endbibitem

%%% 49
\bibitem[\protect\citeauthoryear{Liao et~al.}{2025}]{liao2025memorysammemorizemodalitiessemantics}
\begin{botherref}
\oauthor{\bsnm{Liao}, \binits{C.}},
\oauthor{\bsnm{Zheng}, \binits{X.}},
\oauthor{\bsnm{Lyu}, \binits{Y.}},
\oauthor{\bsnm{Xue}, \binits{H.}},
\oauthor{\bsnm{Cao}, \binits{Y.}},
\oauthor{\bsnm{Wang}, \binits{J.}},
\oauthor{\bsnm{Yang}, \binits{K.}},
\oauthor{\bsnm{Hu}, \binits{X.}}:
Memorysam: Memorize modalities and semantics with segment anything model 2 for multi-modal semantic segmentation.
arXiv preprint arXiv:2503.06700
(2025)
\end{botherref}
\endbibitem

%%% 50
\bibitem[\protect\citeauthoryear{Zhao et~al.}{2025}]{zhao2025eseg}
\begin{bchapter}
\bauthor{\bsnm{Zhao}, \binits{Y.}},
\bauthor{\bsnm{Lyu}, \binits{G.}},
\bauthor{\bsnm{Li}, \binits{K.}},
\bauthor{\bsnm{Wang}, \binits{Z.}},
\bauthor{\bsnm{Chen}, \binits{H.}},
\bauthor{\bsnm{Yang}, \binits{Z.}},
\bauthor{\bsnm{Deng}, \binits{Y.}}:
\bctitle{{ESEG:} event-based segmentation boosted by explicit edge-semantic guidance}.
In: \beditor{\bsnm{Walsh}, \binits{T.}},
\beditor{\bsnm{Shah}, \binits{J.}},
\beditor{\bsnm{Kolter}, \binits{Z.}} (eds.)
\bbtitle{Association for the Advancement of Artificial Intelligence (AAAI)},
pp. \bfpage{10510}--\blpage{10518}
(\byear{2025})
\end{bchapter}
\endbibitem

%%% 51
\bibitem[\protect\citeauthoryear{Li et~al.}{2025}]{li2025know}
\begin{bchapter}
\bauthor{\bsnm{Li}, \binits{K.}},
\bauthor{\bsnm{Lyu}, \binits{G.}},
\bauthor{\bsnm{Chen}, \binits{H.}},
\bauthor{\bsnm{Xie}, \binits{B.}},
\bauthor{\bsnm{Yang}, \binits{Z.}},
\bauthor{\bsnm{Li}, \binits{Y.}},
\bauthor{\bsnm{Deng}, \binits{Y.}}:
\bctitle{Know where you are from: Event-based segmentation via spatio-temporal propagation}.
In: \beditor{\bsnm{Walsh}, \binits{T.}},
\beditor{\bsnm{Shah}, \binits{J.}},
\beditor{\bsnm{Kolter}, \binits{Z.}} (eds.)
\bbtitle{Association for the Advancement of Artificial Intelligence (AAAI)},
pp. \bfpage{4806}--\blpage{4814}
(\byear{2025})
\end{bchapter}
\endbibitem

%%% 52
\bibitem[\protect\citeauthoryear{Kong et~al.}{2024}]{kong2024openess}
\begin{bchapter}
\bauthor{\bsnm{Kong}, \binits{L.}},
\bauthor{\bsnm{Liu}, \binits{Y.}},
\bauthor{\bsnm{Ng}, \binits{L.X.}},
\bauthor{\bsnm{Cottereau}, \binits{B.R.}},
\bauthor{\bsnm{Ooi}, \binits{W.T.}}:
\bctitle{Openess: Event-based semantic scene understanding with open vocabularies}.
In: \bbtitle{{IEEE/CVF} Conference on Computer Vision and Pattern Recognition (CVPR)},
pp. \bfpage{15686}--\blpage{15698}
(\byear{2024})
\end{bchapter}
\endbibitem

%%% 53
\bibitem[\protect\citeauthoryear{Rebecq et~al.}{2019}]{rebecq2019events}
\begin{bchapter}
\bauthor{\bsnm{Rebecq}, \binits{H.}},
\bauthor{\bsnm{Ranftl}, \binits{R.}},
\bauthor{\bsnm{Koltun}, \binits{V.}},
\bauthor{\bsnm{Scaramuzza}, \binits{D.}}:
\bctitle{Events-to-video: Bringing modern computer vision to event cameras}.
In: \bbtitle{{IEEE} Conference on Computer Vision and Pattern Recognition (CVPR)},
pp. \bfpage{3857}--\blpage{3866}
(\byear{2019})
\end{bchapter}
\endbibitem

%%% 54
\bibitem[\protect\citeauthoryear{Gehrig et~al.}{2020}]{gehrig2020video}
\begin{bchapter}
\bauthor{\bsnm{Gehrig}, \binits{D.}},
\bauthor{\bsnm{Gehrig}, \binits{M.}},
\bauthor{\bsnm{Hidalgo{-}Carri{\'{o}}}, \binits{J.}},
\bauthor{\bsnm{Scaramuzza}, \binits{D.}}:
\bctitle{Video to events: Recycling video datasets for event cameras}.
In: \bbtitle{2020 {IEEE/CVF} Conference on Computer Vision and Pattern Recognition (CVPR)},
pp. \bfpage{3583}--\blpage{3592}
(\byear{2020})
\end{bchapter}
\endbibitem

%%% 55
\bibitem[\protect\citeauthoryear{Wang et~al.}{2021a}]{wang2021evdistill}
\begin{bchapter}
\bauthor{\bsnm{Wang}, \binits{L.}},
\bauthor{\bsnm{Chae}, \binits{Y.}},
\bauthor{\bsnm{Yoon}, \binits{S.}},
\bauthor{\bsnm{Kim}, \binits{T.}},
\bauthor{\bsnm{Yoon}, \binits{K.}}:
\bctitle{Evdistill: Asynchronous events to end-task learning via bidirectional reconstruction-guided cross-modal knowledge distillation}.
In: \bbtitle{{IEEE} Conference on Computer Vision and Pattern Recognition (CVPR)},
pp. \bfpage{608}--\blpage{619}
(\byear{2021})
\end{bchapter}
\endbibitem

%%% 56
\bibitem[\protect\citeauthoryear{Wang et~al.}{2021b}]{wang2021dual}
\begin{bchapter}
\bauthor{\bsnm{Wang}, \binits{L.}},
\bauthor{\bsnm{Chae}, \binits{Y.}},
\bauthor{\bsnm{Yoon}, \binits{K.}}:
\bctitle{Dual transfer learning for event-based end-task prediction via pluggable event to image translation}.
In: \bbtitle{2021 {IEEE/CVF} International Conference on Computer Vision (ICCV)},
pp. \bfpage{2115}--\blpage{2125}
(\byear{2021})
\end{bchapter}
\endbibitem

%%% 57
\bibitem[\protect\citeauthoryear{Sun et~al.}{2022}]{sun2022ess}
\begin{bchapter}
\bauthor{\bsnm{Sun}, \binits{Z.}},
\bauthor{\bsnm{Messikommer}, \binits{N.}},
\bauthor{\bsnm{Gehrig}, \binits{D.}},
\bauthor{\bsnm{Scaramuzza}, \binits{D.}}:
\bctitle{{ESS:} learning event-based semantic segmentation from still images}.
In: \beditor{\bsnm{Avidan}, \binits{S.}},
\beditor{\bsnm{Brostow}, \binits{G.J.}},
\beditor{\bsnm{Ciss{\'{e}}}, \binits{M.}},
\beditor{\bsnm{Farinella}, \binits{G.M.}},
\beditor{\bsnm{Hassner}, \binits{T.}} (eds.)
\bbtitle{European Conference on Computer Vision (ECCV)},
vol. \bseriesno{13694},
pp. \bfpage{341}--\blpage{357}.
\bpublisher{Springer}, \blocation{???}
(\byear{2022})
\end{bchapter}
\endbibitem

%%% 58
\bibitem[\protect\citeauthoryear{Jia et~al.}{2023}]{jia2023event}
\begin{barticle}
\bauthor{\bsnm{Jia}, \binits{Z.}},
\bauthor{\bsnm{You}, \binits{K.}},
\bauthor{\bsnm{He}, \binits{W.}},
\bauthor{\bsnm{Tian}, \binits{Y.}},
\bauthor{\bsnm{Feng}, \binits{Y.}},
\bauthor{\bsnm{Wang}, \binits{Y.}},
\bauthor{\bsnm{Jia}, \binits{X.}},
\bauthor{\bsnm{Lou}, \binits{Y.}},
\bauthor{\bsnm{Zhang}, \binits{J.}},
\bauthor{\bsnm{Li}, \binits{G.}},
\bauthor{\bsnm{Zhang}, \binits{Z.}}:
\batitle{Event-based semantic segmentation with posterior attention}.
\bjtitle{{IEEE} Trans. Image Process.}
\bvolume{32},
\bfpage{1829}--\blpage{1842}
(\byear{2023})
\end{barticle}
\endbibitem

%%% 59
\bibitem[\protect\citeauthoryear{Zhang et~al.}{2022}]{zhang2021exploring}
\begin{barticle}
\bauthor{\bsnm{Zhang}, \binits{J.}},
\bauthor{\bsnm{Yang}, \binits{K.}},
\bauthor{\bsnm{Stiefelhagen}, \binits{R.}}:
\batitle{Exploring event-driven dynamic context for accident scene segmentation}.
\bjtitle{{IEEE} Trans. Intell. Transp. Syst.}
\bvolume{23}(\bissue{3}),
\bfpage{2606}--\blpage{2622}
(\byear{2022})
\end{barticle}
\endbibitem

%%% 60
\bibitem[\protect\citeauthoryear{Messikommer et~al.}{2022}]{messikommer2022bridging}
\begin{barticle}
\bauthor{\bsnm{Messikommer}, \binits{N.}},
\bauthor{\bsnm{Gehrig}, \binits{D.}},
\bauthor{\bsnm{Gehrig}, \binits{M.}},
\bauthor{\bsnm{Scaramuzza}, \binits{D.}}:
\batitle{Bridging the gap between events and frames through unsupervised domain adaptation}.
\bjtitle{{IEEE} Robotics Autom. Lett.}
\bvolume{7}(\bissue{2}),
\bfpage{3515}--\blpage{3522}
(\byear{2022})
\end{barticle}
\endbibitem

%%% 61
\bibitem[\protect\citeauthoryear{Hamaguchi et~al.}{2023}]{hamaguchi2023hierarchical}
\begin{bchapter}
\bauthor{\bsnm{Hamaguchi}, \binits{R.}},
\bauthor{\bsnm{Furukawa}, \binits{Y.}},
\bauthor{\bsnm{Onishi}, \binits{M.}},
\bauthor{\bsnm{Sakurada}, \binits{K.}}:
\bctitle{Hierarchical neural memory network for low latency event processing}.
In: \bbtitle{{IEEE/CVF} Conference on Computer Vision and Pattern Recognition (CVPR)},
pp. \bfpage{22867}--\blpage{22876}
(\byear{2023})
\end{bchapter}
\endbibitem

%%% 62
\bibitem[\protect\citeauthoryear{Biswas et~al.}{2024}]{das2024halsie}
\begin{bchapter}
\bauthor{\bsnm{Biswas}, \binits{S.D.}},
\bauthor{\bsnm{Kosta}, \binits{A.}},
\bauthor{\bsnm{Liyanagedera}, \binits{C.M.}},
\bauthor{\bsnm{Apolinario}, \binits{M.P.E.}},
\bauthor{\bsnm{Roy}, \binits{K.}}:
\bctitle{{HALSIE:} hybrid approach to learning segmentation by simultaneously exploiting image and event modalities}.
In: \bbtitle{{IEEE/CVF} Winter Conference on Applications of Computer Vision (WACV)},
pp. \bfpage{5952}--\blpage{5962}.
\bpublisher{{IEEE}}, \blocation{???}
(\byear{2024})
\end{bchapter}
\endbibitem

%%% 63
\bibitem[\protect\citeauthoryear{Liu et~al.}{2025}]{liu2025sharecmp}
\begin{barticle}
\bauthor{\bsnm{Liu}, \binits{Z.}},
\bauthor{\bsnm{Wang}, \binits{B.}},
\bauthor{\bsnm{Wang}, \binits{L.}},
\bauthor{\bsnm{Mao}, \binits{C.}},
\bauthor{\bsnm{Li}, \binits{Y.}}:
\batitle{Sharecmp: Polarization-aware {RGB-P} semantic segmentation}.
\bjtitle{{IEEE} Trans. Circuits Syst. Video Technol.}
\bvolume{35}(\bissue{10}),
\bfpage{10316}--\blpage{10329}
(\byear{2025})
\end{barticle}
\endbibitem

%%% 64
\bibitem[\protect\citeauthoryear{Chen et~al.}{2021}]{chen2021dynamic}
\begin{bchapter}
\bauthor{\bsnm{Chen}, \binits{J.}},
\bauthor{\bsnm{Wang}, \binits{X.}},
\bauthor{\bsnm{Guo}, \binits{Z.}},
\bauthor{\bsnm{Zhang}, \binits{X.}},
\bauthor{\bsnm{Sun}, \binits{J.}}:
\bctitle{Dynamic region-aware convolution}.
In: \bbtitle{{IEEE} Conference on Computer Vision and Pattern Recognition (CVPR)},
pp. \bfpage{8064}--\blpage{8073}
(\byear{2021})
\end{bchapter}
\endbibitem

%%% 65
\bibitem[\protect\citeauthoryear{Zhou et~al.}{2021}]{zhou2021decoupled}
\begin{bchapter}
\bauthor{\bsnm{Zhou}, \binits{J.}},
\bauthor{\bsnm{Jampani}, \binits{V.}},
\bauthor{\bsnm{Pi}, \binits{Z.}},
\bauthor{\bsnm{Liu}, \binits{Q.}},
\bauthor{\bsnm{Yang}, \binits{M.}}:
\bctitle{Decoupled dynamic filter networks}.
In: \bbtitle{{IEEE} Conference on Computer Vision and Pattern Recognition (CVPR)},
pp. \bfpage{6647}--\blpage{6656}
(\byear{2021})
\end{bchapter}
\endbibitem

%%% 66
\bibitem[\protect\citeauthoryear{Li et~al.}{2025}]{li2025u3m}
\begin{barticle}
\bauthor{\bsnm{Li}, \binits{B.}},
\bauthor{\bsnm{Zhang}, \binits{D.}},
\bauthor{\bsnm{Zhao}, \binits{Z.}},
\bauthor{\bsnm{Gao}, \binits{J.}},
\bauthor{\bsnm{Li}, \binits{X.}}:
\batitle{{U3M:} unbiased multiscale modal fusion model for multimodal semantic segmentation}.
\bjtitle{Pattern Recognit.}
\bvolume{168},
\bfpage{111801}
(\byear{2025})
\end{barticle}
\endbibitem

%%% 67
\bibitem[\protect\citeauthoryear{Orsic et~al.}{2019}]{orsic2019defense}
\begin{bchapter}
\bauthor{\bsnm{Orsic}, \binits{M.}},
\bauthor{\bsnm{Kreso}, \binits{I.}},
\bauthor{\bsnm{Bevandic}, \binits{P.}},
\bauthor{\bsnm{Segvic}, \binits{S.}}:
\bctitle{In defense of pre-trained imagenet architectures for real-time semantic segmentation of road-driving images}.
In: \bbtitle{{IEEE} Conference on Computer Vision and Pattern Recognition (CVPR)},
pp. \bfpage{12607}--\blpage{12616}
(\byear{2019})
\end{bchapter}
\endbibitem

%%% 68
\bibitem[\protect\citeauthoryear{Wang et~al.}{2024}]{wang2024multisenseseg}
\begin{barticle}
\bauthor{\bsnm{Wang}, \binits{Q.}},
\bauthor{\bsnm{Chen}, \binits{W.}},
\bauthor{\bsnm{Huang}, \binits{Z.}},
\bauthor{\bsnm{Tang}, \binits{H.}},
\bauthor{\bsnm{Yang}, \binits{L.}}:
\batitle{Multisenseseg: {A} cost-effective unified multimodal semantic segmentation model for remote sensing}.
\bjtitle{{IEEE} Trans. Geosci. Remote. Sens.}
\bvolume{62},
\bfpage{1}--\blpage{24}
(\byear{2024})
\end{barticle}
\endbibitem

%%% 69
\bibitem[\protect\citeauthoryear{Lu et~al.}{2025}]{lu2024polarization}
\begin{barticle}
\bauthor{\bsnm{Lu}, \binits{Y.}},
\bauthor{\bsnm{Tian}, \binits{J.}},
\bauthor{\bsnm{Su}, \binits{Y.}},
\bauthor{\bsnm{Liu}, \binits{J.}},
\bauthor{\bsnm{Hao}, \binits{C.}},
\bauthor{\bsnm{Yue}, \binits{C.}},
\bauthor{\bsnm{Yang}, \binits{X.}}:
\batitle{Polarization-aware low-light image enhancement for nighttime intelligent vehicles}.
\bjtitle{{IEEE} Trans. Intell. Veh.}
\bvolume{10}(\bissue{3}),
\bfpage{1944}--\blpage{1958}
(\byear{2025})
\end{barticle}
\endbibitem

%%% 70
\bibitem[\protect\citeauthoryear{Yan et~al.}{2021}]{yan2021nlfnet}
\begin{bchapter}
\bauthor{\bsnm{Yan}, \binits{R.}},
\bauthor{\bsnm{Yang}, \binits{K.}},
\bauthor{\bsnm{Wang}, \binits{K.}}:
\bctitle{Nlfnet: Non-local fusion towards generalized multimodal semantic segmentation across rgb-depth, polarization, and thermal images}.
In: \bbtitle{{IEEE} International Conference on Robotics and Biomimetics (ROBIO)},
pp. \bfpage{1129}--\blpage{1135}
(\byear{2021})
\end{bchapter}
\endbibitem

%%% 71
\bibitem[\protect\citeauthoryear{Huang et~al.}{2025}]{Huang2024RoadFormerDR}
\begin{barticle}
\bauthor{\bsnm{Huang}, \binits{J.}},
\bauthor{\bsnm{Li}, \binits{J.}},
\bauthor{\bsnm{Jia}, \binits{N.}},
\bauthor{\bsnm{Sun}, \binits{Y.}},
\bauthor{\bsnm{Liu}, \binits{C.}},
\bauthor{\bsnm{Chen}, \binits{Q.}},
\bauthor{\bsnm{Fan}, \binits{R.}}:
\batitle{Roadformer+: Delivering {RGB-X} scene parsing through scale-aware information decoupling and advanced heterogeneous feature fusion}.
\bjtitle{{IEEE} Trans. Intell. Veh.}
\bvolume{10}(\bissue{5}),
\bfpage{3156}--\blpage{3165}
(\byear{2025})
\end{barticle}
\endbibitem

%%% 72
\bibitem[\protect\citeauthoryear{Zhao et~al.}{2017}]{zhao2017pyramid}
\begin{bchapter}
\bauthor{\bsnm{Zhao}, \binits{H.}},
\bauthor{\bsnm{Shi}, \binits{J.}},
\bauthor{\bsnm{Qi}, \binits{X.}},
\bauthor{\bsnm{Wang}, \binits{X.}},
\bauthor{\bsnm{Jia}, \binits{J.}}:
\bctitle{Pyramid scene parsing network}.
In: \bbtitle{2017 {IEEE} Conference on Computer Vision and Pattern Recognition (CVPR)},
pp. \bfpage{6230}--\blpage{6239}
(\byear{2017})
\end{bchapter}
\endbibitem

%%% 73
\bibitem[\protect\citeauthoryear{Yuan et~al.}{2020}]{yuan2020object}
\begin{bchapter}
\bauthor{\bsnm{Yuan}, \binits{Y.}},
\bauthor{\bsnm{Chen}, \binits{X.}},
\bauthor{\bsnm{Wang}, \binits{J.}}:
\bctitle{Object-contextual representations for semantic segmentation}.
In: \bbtitle{European Conference on Computer Vision (ECCV)},
vol. \bseriesno{12351},
pp. \bfpage{173}--\blpage{190}
(\byear{2020})
\end{bchapter}
\endbibitem

%%% 74
\bibitem[\protect\citeauthoryear{Zhuang et~al.}{2021}]{zhuang2020video}
\begin{barticle}
\bauthor{\bsnm{Zhuang}, \binits{J.}},
\bauthor{\bsnm{Wang}, \binits{Z.}},
\bauthor{\bsnm{Wang}, \binits{B.}}:
\batitle{Video semantic segmentation with distortion-aware feature correction}.
\bjtitle{{IEEE} Trans. Circuits Syst. Video Technol.}
\bvolume{31}(\bissue{8}),
\bfpage{3128}--\blpage{3139}
(\byear{2021})
\end{barticle}
\endbibitem

%%% 75
\bibitem[\protect\citeauthoryear{Wang et~al.}{2021}]{wang2021temporal}
\begin{bchapter}
\bauthor{\bsnm{Wang}, \binits{H.}},
\bauthor{\bsnm{Wang}, \binits{W.}},
\bauthor{\bsnm{Liu}, \binits{J.}}:
\bctitle{Temporal memory attention for video semantic segmentation}.
In: \bbtitle{2021 {IEEE} International Conference on Image Processing (ICIP)},
pp. \bfpage{2254}--\blpage{2258}
(\byear{2021})
\end{bchapter}
\endbibitem

%%% 76
\bibitem[\protect\citeauthoryear{Seichter et~al.}{2021}]{seichter2021efficient}
\begin{bchapter}
\bauthor{\bsnm{Seichter}, \binits{D.}},
\bauthor{\bsnm{K{\"{o}}hler}, \binits{M.}},
\bauthor{\bsnm{Lewandowski}, \binits{B.}},
\bauthor{\bsnm{Wengefeld}, \binits{T.}},
\bauthor{\bsnm{Gross}, \binits{H.}}:
\bctitle{Efficient {RGB-D} semantic segmentation for indoor scene analysis}.
In: \bbtitle{{IEEE} International Conference on Robotics and Automation (ICRA)},
pp. \bfpage{13525}--\blpage{13531}
(\byear{2021})
\end{bchapter}
\endbibitem

%%% 77
\bibitem[\protect\citeauthoryear{Chen et~al.}{2020}]{chen2020bi}
\begin{bchapter}
\bauthor{\bsnm{Chen}, \binits{X.}},
\bauthor{\bsnm{Lin}, \binits{K.-Y.}},
\bauthor{\bsnm{Wang}, \binits{J.}},
\bauthor{\bsnm{Wu}, \binits{W.}},
\bauthor{\bsnm{Qian}, \binits{C.}},
\bauthor{\bsnm{Li}, \binits{H.}},
\bauthor{\bsnm{Zeng}, \binits{G.}}:
\bctitle{Bi-directional cross-modality feature propagation with separation-and-aggregation gate for rgb-d semantic segmentation}.
In: \bbtitle{European Conference on Computer Vision (ECCV)}
(\byear{2020})
\end{bchapter}
\endbibitem

\end{thebibliography}
